\documentclass[runningheads]{llncs}

\usepackage{graphicx}
\usepackage{tikz}
\usepackage{comment}
\usepackage{amsmath,amssymb} 
\usepackage{color}

\usepackage[accsupp]{axessibility}  

\usepackage[width=122mm,left=12mm,paperwidth=146mm,height=193mm,top=12mm,paperheight=217mm]{geometry}

\usepackage{xspace}

\makeatletter
\DeclareRobustCommand\onedot{\futurelet\@let@token\@onedot}
\def\@onedot{\ifx\@let@token.\else.\null\fi\xspace}

\def\ie{\emph{i.e}\onedot}

\makeatother

\usepackage{graphicx}
\usepackage{amssymb}
\usepackage{booktabs}
\usepackage{xcolor}
\usepackage{wrapfig}
\usepackage{paralist}
\usepackage[pagebackref,breaklinks,colorlinks]{hyperref}
\definecolor{RowColorCode}{rgb}{0.61,0.57,0.89}
\usepackage{color, colortbl}
\usepackage{amssymb}
\usepackage{subcaption}
\usepackage{wrapfig}

\usepackage{pifont}
\newcommand{\cmark}{\ding{51}}
\newcommand{\xmark}{\ding{55}}

\usepackage[capitalize]{cleveref}
\crefname{section}{Sec.}{Secs.}
\Crefname{section}{Section}{Sections}
\Crefname{table}{Table}{Tables}
\crefname{table}{Tab.}{Tabs.}

\begin{document}

\newcommand{\bigO}[1]{\mathcal{O}(#1)}
\newcommand{\sota}{state-of-the-art}
\newcommand{\tian}[1]{\textcolor{orange}{#1}}
\newcommand{\divya}[1]{\textcolor{blue}{#1}}
\newcommand{\model}{FAR}
\newcommand{\datanames}{Sub-$k$ matrices}
\newcommand{\G}{$G$}
\newcommand{\V}{$V$}
\newcommand{\E}{$E$}
\newcommand{\brr}[1]{\left( #1 \right)}
\newcommand{\bcc}[1]{ \left{ #1 \right} }
\newcommand{\bss}[1]{\left[ #1 \right]}
\newcommand{\mc}[1]{\mathcal{#1}}
\newcommand{\sg}{\mathcal{L}}
\newcommand{\li}{\sg}
\newcommand{\vts}[1]{\lvert #1 \rvert}
\newcommand{\Vts}[1]{\lVert #1 \rVert}
\newcommand{\bb}[1]{\mathbb{#1}}
\newcommand\inv[1]{#1\raisebox{1.05ex}{$\scriptscriptstyle-\!1$}}
\newcommand\Tstrut{\rule{0pt}{2.6ex}}         
\newcommand\Bstrut{\rule[-1.3ex]{0pt}{0pt}}   
\newcommand\Bstrutfrac{\rule[-0.7ex]{0pt}{0pt}}   
\newcommand\Tstrutfrac{\rule{0pt}{1.7ex}}         
\newcommand\mathdash{\text{\normalfont --}}
\newcommand{\cost}{\bigO{ \vts{\li^{\scriptscriptstyle -1}_t}k }}
\newcommand{\cm}{\mathcal{M}_{\Delta t}(u)}
\newcommand{\pc}{\zeta_c(t)}
\newcommand{\pd}{\zeta_d(t)}
\newcommand{\pe}{\zeta_e(t)}
\newcommand{\costk}{\bigO{ \vts{\li^{\scriptscriptstyle -1}_t} }}
\newcommand\setrow[1]{\gdef\rowmac{#1}#1\ignorespaces}
\makeatletter
\newcommand\footnoteref[1]{\protected@xdef\@thefnmark{\ref{#1}}\@footnotemark}
\makeatother
\newcommand{\size}{\bigO{d}}
\newcommand{\shorteq}{%
  \settowidth{\@tempdima}{-}
  \resizebox{\@tempdima}{\height}{=}%
}
\newcommand*\midpoint[1]{\overline{#1}}
\newcommand{\mysetminus}{\mathbin{\fgebackslash}}


\let\proof\relax
\let\endproof\relax

\linespread{0.97}
\setlength{\parskip}{-0.1em}
\mathchardef\mhyphen="2D

\newcommand{\minus}{\scalebox{0.75}[1.0]{$-$}}



\title{FAR: Fourier Aerial Video Recognition}


\author{Divya Kothandaraman\inst{1}, Tianrui Guan\inst{1}, Xijun Wang\inst{1}, Sean Hu\inst{2}, Ming Lin\inst{1}, Dinesh Manocha\inst{1}}
\institute{University of Maryland College Park, United States \and
DEVCOM Army Research Laboratory, United States \\
\email{dkr@umd.edu}\\
\url{https://gamma.umd.edu/far} }

\titlerunning{} 
\authorrunning{} 

\maketitle

\begin{abstract}
   We present an algorithm, Fourier Activity Recognition (FAR), for UAV video activity recognition. Our formulation uses a novel Fourier object disentanglement method to innately separate out the human agent (which is typically small) from the  background. Our disentanglement technique operates in the frequency domain to characterize the extent of temporal change of spatial pixels, and exploits convolution-multiplication properties of Fourier transform to map this representation to the corresponding object-background entangled features obtained from the network. To encapsulate contextual information and long-range space-time dependencies, we present a novel Fourier Attention algorithm, which emulates the benefits of self-attention by modeling the weighted outer product in the frequency domain. Our Fourier attention formulation uses much fewer computations than self-attention. 
   We have evaluated our approach on multiple UAV datasets including UAV Human RGB, UAV Human Night, Drone Action, and NEC Drone. We demonstrate a relative improvement of {\bf 8.02\% - 38.69\%} in top-1 accuracy and up to {\bf 3} times faster over prior works.
\end{abstract}

\section{Introduction}

\label{sec:intro}

Deep learning techniques have been widely used for activity recognition \cite{feichtenhofer2020x3d,carreira2017quo,bertasius2021space}. Video analysis of scenes captured using UAV cameras \cite{li2021uav,perera2019drone} is much harder than activity recognition in ground-camera datasets \cite{carreira2017quo,sigurdsson2018charades}. In these UAV videos, the object of interest, \ie the human actor (any individual appearing in the video performing scripted or non-scripted actions), is typically much smaller in terms of number of pixels or the area than the corresponding background, and thus provides less knowledge than a front view capture. Moreover, it is harder to capture and label UAV videos. Overall, there are fewer and smaller labeled datasets of aerial videos, as compared to ground videos. For instance, ground-camera datasets like Kinetics-400~\cite{carreira2017quo} contain $306,245$ videos while the recent  UAV-Human~\cite{li2021uav} database has $22,476$ videos. 
\footnotetext{The second and third authors contributed equally}

Given that the size of the human actor in UAV videos is much smaller than the corresponding background, a neural network trained on these datasets may learn to infer more from the background~\cite{li2018tell} than the  human actor. While both background and context are important \cite{choi2019can}, the network must learn to first identify the human actor and the corresponding action, and then deduce relations of the human actor with the surroundings in a judicious manner. In the absence of annotated detection boxes that can demarcate the human actor, the network needs to be able to differentiate the moving human actor from the background in an intrinsic manner. One approach is to detect the object of interest via object detection \cite{ren2015faster}. However, action recognition models that heavily rely on localization of the human actor require near to perfect object detection accuracy~\cite{zou2019object}. While it is practically not feasible to annotate all datasets for object detection, object detectors trained on ground camera datasets will not generalize well UAV videos due to domain gap issues~\cite{wang2018deep,chen2018domain,benjdira2019unsupervised}. Domain adaptation solutions do not lead to perfect generalization yet. 

On the other hand, traditional optical flow \cite{beauchemin1995computation} techniques require hundreds of optimization iterations each frame, and split the network into RGB and motion streams which increases computation and model parameters \cite{piergiovanni2019representation}. Low computation alternatives such as deep learning based optical flow \cite{ren2017unsupervised,ilg2017flownet,dosovitskiy2015flownet}, motion feature networks \cite{lee2018motion} and ActionFlowNet \cite{ng2018actionflownet} are inferior in performance compared to optical flow. Techniques such as background subtraction \cite{piccardi2004background} and motion segmentation \cite{zappella2008motion} are not very promising either \cite{sengupta2020background,ellenfeld2021deep}. Thus, the network needs to learn to automatically \textit{disentangle}~\cite{zhu2020object,dundar2020unsupervised} object feature representations from the corresponding entangled state containing both the object and background information.  

In addition to object-background separation, it is important for the network to acquire knowledge \cite{bertasius2021space} about the context, relationships between the object, and background and intra-pixels correspondences as well. Self-attention~\cite{vaswani2017attention,zhang2019self} can model this information by capturing long-range dependencies within an image/ video. Prior work on attention based video activity recognition~\cite{bertasius2021space,liu2021video,arnab2021vivit} has seen two classes of self attention networks by either directly applying self-attention on convolutional layers or using self-attention as the building block. Mathematically, the core step in the computation of self-attention is matrix multiplication, which makes it computationally expensive. 

\subsection{Main contributions}
We present a novel method, \model~, for UAV video action recognition.  The design of \model~in the frequency domain is motivated by the fact that frequency spectrums contain knowledge about a signals' characteristics that are not easily interpretable in the time domain. Our novel components include:
\begin{itemize}
    \item We propose a novel Fourier Object Disentanglement method (FO) to bestow the network with the ability to \textit{intrinsically} recognize the moving human actor from the background. FO operates in the frequency domain dictated by the spectrum of the Fourier transform corresponding to the temporal dimensions of the video. 
    It characterizes the motion of the human actor based on the magnitude and rate of temporal change of feature maps that encode information about the spatial pixels of the video. The amplitudes at each spatial-temporal location of the feature maps are innately representative of dynamic salient, static salient, dynamic non-salient and static non-salient regions, in the same order of relevance. This also empowers the network to handle videos with moving background pixels and dynamic cameras.
    \item We present Fourier Attention (FA) to encapsulate context and long range space-time dependencies within a video. Fourier attention works in the frequency domain corresponding to the space-time dimensions of the video, and emulates the benefits of self-attention. The time complexity of FA is $O(n^{2}logn)$ as opposed to $O({n^3})$ for traditional self-attention, and the accuracy of Fourier-attention approximates that of self-attention.  
\end{itemize}
 Moreover, such a representation promotes global mixing. \model~has multiple benefits. \begin{inparaenum}[(i)]\item {It elegantly exploits the mathematical properties of Fourier transform to achieve the desired objectives of object background separation and context encoding by performing fewer computations than traditional methods.} \item{It is parameter-less, i.e., it does not have any learnable layers/ parameters.} \item{FAR can be embedded within any 3D action recognition network such as I3D \cite{carreira2017quo,feichtenhofer2020x3d} to achieve state-of-the-art performance.} \item{FAR converges faster than the corresponding 3D action recognition backbone.} \end{inparaenum} 

We experimentally demonstrate that FAR outperforms prior work by $8.02\% - 38.69\%$ performance across multiple UAV datasets including UAV Human RGB~\cite{li2021uav}, UAV Human Night~\cite{li2021uav}, Drone Action~\cite{perera2019drone}, and NEC Drone~\cite{choi2020unsupervised}. We compare with the state-of-the-art Fourier method, efficient attention method and self-attention based transformer methods and demonstrate accuracy, computation and memory benefits. 


\section{Related Work}

\noindent \textbf{Action Recognition:} Action recognition is a well studied topic in computer vision. The emergence of large-scale ground-camera videos datasets~\cite{carreira2017quo,sigurdsson2018charades,monfort2019moments} has led to development of deep learning techniques for action recognition. We refer the reader to ~\cite{actionrecognitionsurvey} for a survey on action recognition. Broadly speaking, three classes of network architectures have been proposed for action recognition. The first ~\cite{simonyan2014two,feichtenhofer2016convolutional,gammulle2017two,hara2018can,wang2019learning} builds on the two-stream theory in cognition to model space and time separately. The second ~\cite{feichtenhofer2019slowfast,feichtenhofer2020x3d,carreira2017quo,bertasius2021space,tran2019video,hussein2019timeception} models space and time jointly via 3D CNNs. The third class includes transformer-based architectures ~\cite{plizzari2020spatial,girdhar2019video,mazzia2021action,bertasius2021space,wang2018non}. These transformer-based solutions are built on self-attention~\cite{vaswani2017attention,zhang2019self} and have high computational complexity. In the interest of optimizing GPU memory, frame sampling strategies~\cite{gowda2020smart,zhi2021mgsampler,korbar2019scsampler,griffin2019bubblenets} for video action recognition have been proposed. The above solutions are focused on challenges pertaining to action recognition in ground-camera videos. However, UAV video action recognition is much more difficult.

\noindent \textbf{UAV Action Recognition:} UAV video databases~\cite{barekatain2017okutama,li2021uav,choi2020unsupervised,perera2019drone} have been used to develop solutions~\cite{sultani2021human,peng2020fully,ding2020lightweight,ulhaq2016action} for UAV action recognition. However, these solutions are directly based off techniques designed for ground-camera datasets~\cite{carreira2017quo,monfort2019moments}, where the size of the object is comparable to the background. Moreover, for ground camera videos, an auxilliary guidance factor based on object detection~\cite{ren2015faster} is a viable option. However, these assumptions do not hold true in UAV videos \cite{du2018unmanned,mittal2020deep}.

\noindent \textbf{FFT and Deep Learning} FFT has been immensely used in traditional image \cite{buijs1974implementation,reddy1996fft} and video processing \cite{zhang2008contour,chun1999automatic} applications. Fast Fourier Transform (FFT) has been recently used in deep learning methods. One of its first applications was to accelerate convolution operations~\cite{fftconv}. Incorporating FFT between NN layers ~\cite{ffc,lfd,ffhtol} instead of CNNs to transform the feature space to the frequency domain, and aid global mixing of knowledge, has been  used to improve accuracy for image classification, detection and ground-camera action recognition. An interesting application of FFT includes image stylization~\cite{fda} as a guiding factor for domain adaptation. Most recently, FFT was used to naively replace self-attention layers \cite{vaswani2017attention} for NLP applications.

\noindent \textbf{Efficient attention} Methods to improve memory efficiency of transformers include modifications in matrix multiplication \cite{shen2021efficient}, low rank approximations \cite{wang2020linformer}, kernel modifications \cite{katharopoulos2020transformers} for linear time complexity \cite{xiong2021nystromformer,schlag2021linear,kim2020fastformers,li2020linear}. While most of these solutions are focused on NLP and image-based computer vision tasks, EA \cite{shen2021efficient} demonstrates results on temporal action localization and STAR \cite{shi2021star} performs skeleton action recognition. The former can be regarded as a localization task w.r.t. the temporal dimension while the latter uses pose information making the task of classification easier. None of these solutions are customized to UAV action recognition which brings forth different challenges. 


\begin{figure*}
    \centering
    \includegraphics[width=\textwidth]{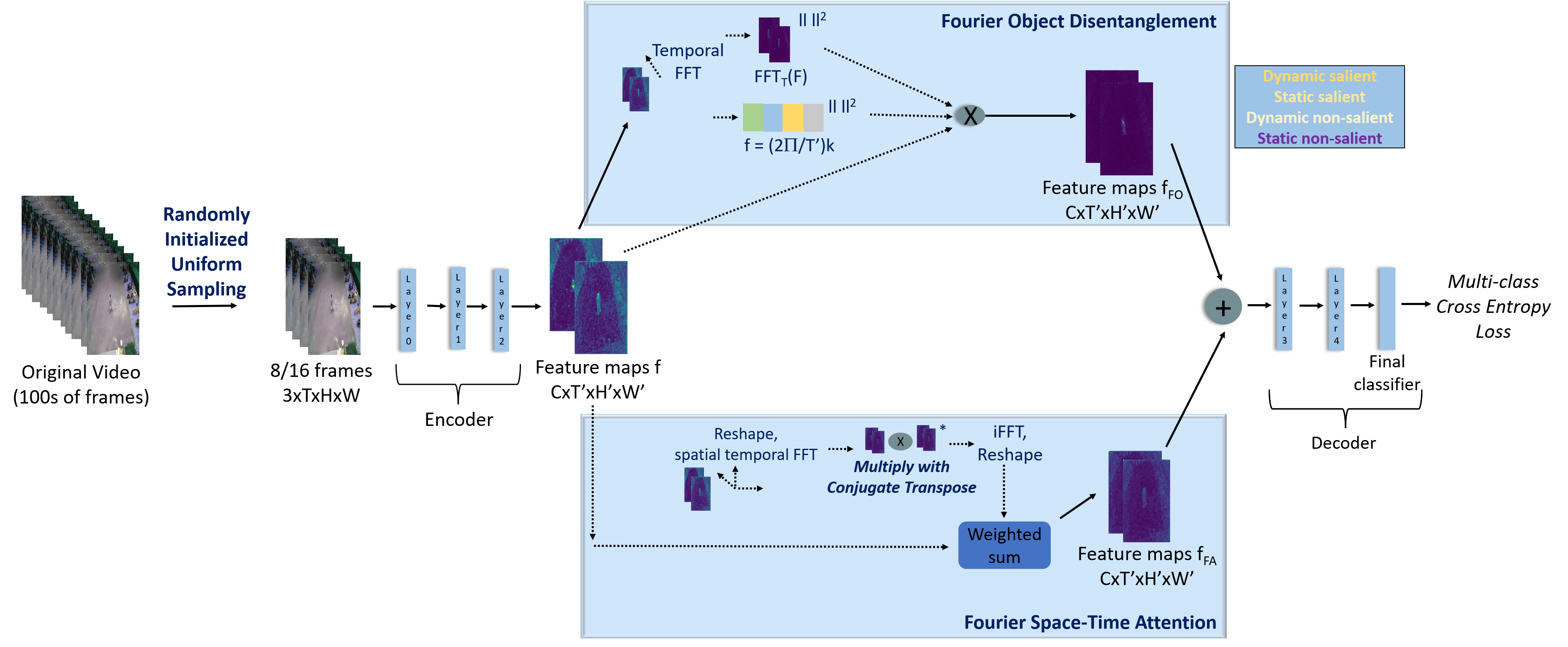}
    \caption{\small{\textbf{Fourier Object Disentanglement (FO) and Space-Time Fourier Attention (FA):} FO empowers the network to intrinsically separate out the moving human agent from the background, without the need for any annotated object detection bounding boxes. This enables our network to explicitly focus on the low resolution human agent performing action, and not just learn from background cues. FO inherently characterizes salient and non-salient, and static and dynamic regions of the scene via the amplitudes of the feature maps it computes. FA elegantly exploits the mathematical properties of the Fourier transform to imbibe the properties of self-attention and capture contextual knowledge and long-range space-time dependencies at a much lower computational complexity.}}
    \label{fig:overview}
    \vspace{-15pt}
\end{figure*}

\section{Fourier Disentangled Space Time Attention}

In this section, we describe our approach. We design two novel methods to decipher the human actor performing action, and encode context. Fourier Object Disentanglement (FO) disentangles the object from the background in an automatic manner. Fourier Space-Time Attention (FA) imbibes the properties of self-attention to capture long range space-time relationships at a lower computational cost. These modules can be embedded within any state-of-the-art 3D video recognition backbone such as I3D \cite{carreira2017quo} or X3D \cite{feichtenhofer2020x3d} for improved action recognition. 
We now describe the methods in detail. 

\subsection{Fourier Object Disentanglement}
\label{method:fo}
We present a Fourier Object Disentanglement (FO) method to automatically separate the human actor from the background. Movement of the human actor in the scene can be characterised by temporal change of feature maps encoding spatial pixels (across space dimensions $H \times W$) in the video frames. The rate and magnitude of change of a signal can be quantified by amplitude of a signal at different frequencies. Thus, to identify the movement, we first transform the feature maps to a temporal frequency space. We perform this computation using 1D Fourier transform along the temporal dimension. Specifically, let $f(c,t,h,w)\in C\times T' \times H' \times W'$ denote the feature maps on which FO is applied, where $C$ is the number of channels and $T'$ and $(H'\times W')$ denote the temporal and spatial dimensions of the feature maps, respectively. The amplitude of the temporal Fourier transform at the frequency $-2\pi k/N$ is: 
\begin{equation}
    \mathcal{F_{T}}(f)(k) = \sum_{n=0}^{n=T'} f(c,t,h,w) \times e^{-2\pi kn/N},
\end{equation}
which can be computed efficiently using the FFT algorithm \cite{frigo1998fftw}. $\mathcal{F_{T}}(f)(k)$ mathematically represents the amplitude of the temporal signal at every spatial and channel location of the feature map $f$, at various frequencies. Intuitively, high frequency in the temporal dimension corresponds to the movement, and low frequency represents static regions of the scene. Therefore, regions corresponding to the moving human actor should have higher amplitude of Fourier transform at high frequencies. To infer the presence of the moving human actor at various spatial locations, we encapsulate the relationships between amplitudes and frequencies by multiplying the L2-norm-square of the amplitude at each frequency with the L2-norm-square of the frequency itself. L2-norm ensures that frequencies and amplitudes are positive. L2-norm-square amplifies high amplitudes of the Fourier transform of the signal at high frequencies and suppress low amplitudes at low frequencies for disentangling dynamic regions of the scene. The frequencies, in order, are:$  fr_{k} = [e^{-2\pi k/N}], k = 1....T'$.
  
Note that the frequencies are independent of the input video. Thus, the dynamic mask  $\mathcal{M}_{FO}$ can be represented as 
\begin{equation}
     M_{FO} = \| {F_{T}}(f)(k) \|_{2}^{2} \times \| fr_{k} \|_{2}^{2},
\end{equation}
where $|a|_{2}^{2}$ is L2 norm-square of a vector $|a|$. $M_{FO}$ disentangles (or amplifies) parts of the scene corresponding to moving pixels. This may include moving background (and camera motion) in addition to moving human actor. Our next task is to use $M_{FO}$ to demarcate moving object pixels from \textbf{moving background pixels}. 

To further separate out only the moving actor, we capitalize on the activation maps $f$ computed by the model. While not perfect, the activations at salient regions of the scene are higher than those at the non-salient regions. Hence, the final object disentangled representations can be represented as a dot product of $M_{FO}$ and network features $f$, which amplifies dynamic, salient regions of the scene. Mathematically,
\begin{equation}
    F_{FO} = f \odot M_{FO}.
    \label{eq:fo_final}
\end{equation}
According to this formulation, dynamic salient regions are amplified the most, and static non-salient regions are heavily suppressed. The amplitude at static salient regions and dynamic non-salient regions is lower than the amplitude at dynamic salient regions. Due to the $l2$ operation in the computation of $M_{FO}$ and linear application of $f$ in Equation \ref{eq:fo_final}, static salient regions have a higher amplitude than the dynamic non-salient regions. Thus, the ordering of amplitudes that is formed as: dynamic-salient $>$ static-salient $>$ dynamic-non-salient $>$ static-non-salient, in concordance with the relevance for decision making for action recognition. Thus, static as well as dynamic background regions have lower amplitudes than static and dynamic regions of the object executing action. 

\noindent \textbf{Time complexity:} The time complexity of FO depends on the time complexity of 1D FFT, which is $nlog(n)$, for an n-element input vector. Consider the classical case \cite{he2016deep} where the temporal and spatial dimensions at the mid level feature representations is half and one-fourth of the number of frames sampled and spatial dimensions of the image respectively. The number of FFTs that need to be computed is $C \times (H/8) \times (W/8)$ where $C, H, W$ correspond to the number of channels at the mid-level, and spatial dimensions of the image. Therefore, the total time complexity is $C \times (H/8) \times (W/8) \times (T/2) log(T/2)$. 

\subsection{Space-Time Fourier Attention}
\label{method:fa}

Consider a scene that depicts a human actor swimming in a swimming pool. Here, it is important to decipher the relationship between the human actor and the pool. While explicit modeling of correspondences between different pixels illustrating  pose, orientation, and joint movements may not be necessary, it is crucial for the neural network to inherently capture this knowledge. Space-time self-attention for video action recognition ~\cite{bertasius2021space,liu2021video,arnab2021vivit} is capable for extracting this knowledge, but comes at the cost for expensive matrix multiplications. 

We propose Fourier Space-Time Attention (FA) for acquiring knowledge about the long-range space-time relations within a video. Fourier attention approximates self-attention  in an elegant fashion at a reduced computational cost. To understand the mechanics of Fourier attention, we first succinctly present self-attention \cite{zhang2019self}. The inputs to self-attention are key, query and value vectors which are representations obtained by $1 \times 1$ convolutions using a common input feature map. Vaswani et al. \cite{vaswani2017attention} describe the computation of self-attention as `` a weighted sum of the values, where the weight (or sub-attention) assigned to each value is computed by a compatibility function of the query with the corresponding key.'' 
Key, query and value are $1\times 1$ convolution layers transforming the input feature maps. Mathematically, with $\mathrm{x}$ representing the input feature maps, and $\odot$ denoting matrix multiplication,
\begin{equation}
    \mathrm{Attention} = \mathrm{Value(x)} \odot [\mathrm{Query(x)}^{T} \odot \mathrm{Key(x)}]^{T}
\end{equation}


Our space-time Fourier attention method proceeds as follows. The first step is to obtain a representation equivalent to the key-query computation, termed \textit{Fourier sub-attention}. Fouurier sub-attention is motivated by autocorrelaton, which is the correlation coefficient between different parts of the same signal. We define \textit{Fourier sub-attention} as the element-wise product of the Fourier transform of feature maps with the conjugate transpose of the Fourier transform of these feature maps (Equation \ref{eq:Fourier_suba}). To compute this space-time Fourier sub-attention, we reshape the video feature maps $f$ to a 3D representation $C\times T' \times (HW)$, which are transformed to the frequency domain via 2D Fourier transform along the space and time axes as follows:
\begin{equation}
    \mathcal{F_{ST}}(f)(m,n) = \sum_{h,w} f(c,t,h,w) e^{-2\pi mh/M} e^{-2\pi nw/N}.
\end{equation}
computed efficiently using the FFT algorithm \cite{frigo1998fftw}. FFT is a representation of the signal as a whole at a wide spectrum of frequencies, and enables inherent and exhaustive global mixing between various spatial and temporal regions of the video. The space-time Fourier sub-attention $\mathcal{A_{ST}}$ in the Fourier domain is simply the element wise multiplication between $\mathcal{F_{ST}}$ and its complex conjugate $\mathcal{F_{ST}}^{*}$:
\begin{equation}
    \mathcal{A_{ST}} = \mathcal{F_{ST}} \times \mathcal{F_{ST}}^{*}
    \label{eq:Fourier_suba}
\end{equation}
Next, we compute the inverse FFT ($\mathcal{IF}$) of $\mathcal{A_{ST}}$ to obtain the correlations in the time domain, and reshape to $C \times T' \times H' \times W'$. These sub-attention ``weights'' are then used in a dot product (or element wise multiplication) with the input feature maps $f$ to compute the final space-time Fourier attention maps $f_{FA}$. A scaling factor $\lambda_{FA}$, chosen empirically to be $0.01$, scales these Fourier attention maps, which are then sum-fused with the input feature maps. Mathematically,
\begin{align}
f_{FA} = F + \lambda_{FA} \times \mathcal{IF}(\mathcal{A_{ST}}), 
\end{align}

\noindent \textbf{Time complexity:} Traditional self-attention \cite{vaswani2017attention} requires the model to perform two matrix multiplications. In the first matrix multiplication of self attention, we multiply the query matrix ($THW \times C$) with the key matrix ($C \times THW$). The time complexity is $O(C \times THW  \times THW)$. In the second matrix multiplication, we multiply the value matrix ($C \times HWT$) with the attention matrix ($HWT \times HWT$). The complexity of this stage is $O(C \times HWT \times HWT)$. Hence, the overall time complexity of space-time self-attention \cite{bertasius2021space} is $O(HWT \times HWT \times C)$. 

In contrast, our Fourier attention solves the problem via one 2D FFT and one 2D iFFT. 2D FFT is computed on a matrix of dimensions $HW \times T$. The number of 2D FFTs that need to be computed is equal to the number of channels ($C$). Hence, the complexity is $O(C \times HWTlog(HWT))$. The complexity of 2D FFT and 2D iFFT are the same. Therefore, the overall time complexity of Fourier attention is $O(C \times HWTog(HWT))$. Clearly, Fourier attention is much more efficient than self attention. In terms of accuracy, space-time Fourier attention is comparable to space-time self-attention \cite{bertasius2021space}.

\subsection{Mathematical Analysis}

\begin{lemma}
Given an input matrix A, Fourier attention as well self-attention \cite{vaswani2017attention,bertasius2021space} encapsulate long-range relationships for global mixing by computing outer products.
\end{lemma}
\paragraph{Proof:} We refer the reader to the supplementary material for the detailed proof. We present a concise version here. Without loss of generality, let $[a_{ij}]$ denote the elements of a square matrix A (with dimensions $N$) in $2D$. $f$, $g$, $h$ represent $1\times 1$ convolutions for key, query, value computations in self-attention. The self-attention matrix $S_{mn}$ is:
\begin{equation}
    S_{mn} = \sum_{l=1}^{N} ha_{ml} \sum_{k=1}^{N} [ ga_{lk} \times fa_{kn}]
    \label{eq:sa}
\end{equation}
Fourier attention $F_{mn}$ is:
\begin{align}
    F_{mn} = \sum_{b=1}^{N} \sum_{c=1}^{N} \overbrace{\exp(\minus 2\pi mc/N) \exp(\minus 2\pi nb/N)}^{h_{mn}(b,c)}a_{mn}  \times \nonumber \\[-10pt] \{\sum_{j=1}^{N} \sum_{i=1}^{N} \underbrace{\exp(\minus 2\pi j(b\minus c)/N)}_{f_{mn}(b,c)} a_{ij} \times \underbrace{\exp(\minus 2\pi i(c\minus b)/N)}_{g_{mn}(b,c)}a_{ij}\}
    \label{eq:fa}
\end{align}

$f$, $g$, $h$ in Equation \ref{eq:sa} are $1\times 1$ convolutions, and that the exponential terms span the entire spectrum of frequencies lets us define $f$, $g$, $h$ for Fourier attention as shown in Equation \ref{eq:fa}. Thus, the equation for Fourier attention can be simplified as:
\begin{equation}
    F_{mn} = \sum_{b=1}^{N} \sum_{c=1}^{N} h_{mn}(b,c)a_{mn} \times \nonumber \{\sum_{j=1}^{N} \sum_{i=1}^{N} f_{mn}(b,c)a_{ij} \times g_{mn}(b,c)a_{ij}\}
    \label{eq:fa1}
\end{equation}
In self-attention, f,g,h are learnable. In contrast, in Fourier attention, f,g,h are pre-defined by the Fourier spectrum. Nonetheless, they exhaustively cover the Fourier spectrum. Moreover, the terms involved and the structure of computations (multiplications followed by summation) in Equations \ref{eq:sa} and \ref{eq:fa1} are similar, both promote global mixing and encapsulate long-range relationships. 

\section{FAR: Activity Recognition in UAVs}

We present FAR, a network for video action recognition in UAVs (Figure \ref{fig:overview}). FAR samples 8-16 frames from the input video by using randomly initialized uniform sampling, described in Section \ref{method:sample}. These frames are passed through the first few layers of the 3D backbone network (or encoder) to generate feature maps $f$. These features contain entangled object and background information along the space-time dimensions. The choice of this intermediate layer in the backbone network that extracts feature maps $f$ is a careful trade-off between the spatial-temporal resolutions needed for FAR to work well and the amount of knowledge contained in the networks' layers. We describe this choice in detail in this section, as well as present ablation experiments in \ref{sec:results_uavhumanrgb} to justify our choice. 

The \textit{Fourier Object Disentanglement} module (Section \ref{method:fo}), and the \textit{Fourier Space-Time Attention} module (Section \ref{method:fa}) act on $f$, in parallel, to generate $f_{FO}$ and $f_{FA}$, respectively. $f_{FO}$ and $f_{FA}$ are sum fused and passed through the remaining layers of the neural network to generate the final action classification probability distribution, used in a multi-class cross entropy loss term with the ground-truth label for back-propagation. 

\noindent \textbf{Incorporating FO within the 3D backbone:} Typically, to encapsulate temporal movement at each spatial location, we need to ensure that the spatial temporal dimensions of the feature map is not too small. Thus, it is useful to perform this operation using mid-level features (output from the middle layer of the network, as shown in Figure \ref{fig:overview}) that strike a fine balance between generic features that capture context, and focused high level features (at output layer).

\noindent \textbf{Incorporating FA within the 3D backbone:} After FO, the network does not contain any background signal. Hence, Fourier attention needs to be applied either before FO or in parallel with FO. FO is applied on mid-level features. Applying FA at a high level is not very effective because the extracted features do not have sufficient information. Hence we apply FA on the mid-level features as well, in parallel with the Fourier object disentanglement module. 

\subsection{Randomly Initialized Uniform Sampling}
\label{method:sample}
It is computationally expensive to use all the frames in a video. In traditional uniform sampling, $T$ frames are sampled at uniform intervals. The standard way of uniform sampling under-utilizes \cite{zhi2021mgsampler,korbar2019scsampler} the knowledge that can be gained from the original video, which adds to the pre-existing issue of limited data. We use a variation of uniform sampling to improve the variance of the network and hence boost accuracy. First, we compute the step size as the ratio of total number of frames in the video and number of frames that we desire to sample. Next, we generate a random number between $0$ and step size, and correspondingly designate the first frame to be sampled. This is followed by uniformly sampling video frames at step size intervals from the designated first frame. 

\section{Experiments and Results}

We will make all code and trained models publicly available. 

\subsection{Datasets}
\label{exp:datasets}
In this section, we briefly describe the UAV datasets used for evaluating FAR. \textbf{UAV Human RGB~\cite{li2021uav}} is the largest UAV-based human behavior understanding dataset. Split $1$ contains $15172$ and $5556$ images for training and testing respectively captured under various adversities including illumination, time of day, weathers, etc. \textbf{UAV Human Night Camera~\cite{li2021uav}} contains videos similar to UAV Human RGB captured using a night-vision camera. The night vision camera captures videos in color mode in the daytime, and grey-scale mode in the nighttime. \textbf{Drone Action~\cite{perera2019drone}} is an outdoor drone video dataset captured using a free flying drone. It has $240$ HD RGB videos across $13$ human actions. \textbf{NEC Drone~\cite{choi2020unsupervised}} is an indoor UAV video dataset with $16$ human actions, performed by human subjects in an unconstrained manner. 

\subsection{Implementation Details}
\label{exp:implementation}
\noindent \textbf{Backbone network architecture:} We benchmark our models using two state-of-the-art video recognition backbone architectures (i) I3D \cite{carreira2017quo} (CVPR 2017) (ii) X3D-M \cite{feichtenhofer2020x3d} (CVPR 2020). 
For both X3D and I3D, we extract mid-level features after the second layer.

\noindent \textbf{Training details:} Our models were trained using NVIDIA GeForce 1080 Ti GPUs, and NVIDIA RTX A5000 GPUs. Initial learning rates were \{$0.01$, and $0.001$\} across datasets. 
We use the Stochastic Gradient Descent (SGD) optimizer with weight decay of $0.0005$ and momentum of $0.9$, and cosine/ poly annealing for learning rate decay. The final softmax predictions of all our models were constrained using multi-class cross entropy loss. 

\noindent \textbf{Evaluation:} We report top-1 and top-5 accuracies. 

\begin{table}
\vspace{-20pt}
\caption{\small{\textbf{Results on UAV Human RGB.} \textbf{Table (a)}: FAR can be embedded within any 3D action recognition backbone to achieve state-of-the-art performance. Pretraining with Kinetics boosts performance, and large input sizes work better since FA and FO are designed to capture global, as well as local knowledge. FAR imparts improvements of $2.20\%$-$38.69\%$ over 3D action recognition backbones across training configurations. \textbf{Table (b) - Ablation experiments}: We demonstrate that each component of FAR imparts substantial improvement in top-1 accuracy by upto {\bf 8\%}.}}
\begin{subtable}{0.6\textwidth}
\centering
\resizebox{0.99\columnwidth}{!}{
\begin{tabular}{c c c c c c c}
\toprule
Backbone & \model~ & Frames & Input Size & Init. & Top-1 & Top-5   \\
\midrule
(i) I3D & \xmark & $8$ & $540\times960$ & Kinetics & $21.06$ & $40.81$ \\
\rowcolor{RowColorCode}
(ii) I3D & \cmark & $8$ & $540\times960$ & Kinetics & $29.21$ & $50.27$\\
\midrule 
(iii) X3D-M & \xmark & $16$ & $224\times224$ & None & $27.0$ & $44.2$ \\
\rowcolor{RowColorCode}
(iv) X3D-M & \cmark & $16$ & $224\times224$ & None & $27.6$ & $44.1$ \\
\midrule 
(v) X3D-M & \xmark & $16$ & $224\times224$ & Kinetics & $30.6$ & $50.3$ \\
\rowcolor{RowColorCode}
(vi) X3D-M & \cmark & $16$ & $224\times224$ & Kinetics & $31.9$ & $50.3$ \\
\midrule 
(vii) X3D-M & \xmark & $8$ & $540\times540$ & Kinetics & $36.6$ & $57.1$ \\
\rowcolor{RowColorCode}
(viii) X3D-M & \cmark & $8$ & $540\times540$ & Kinetics & $38.6$ & $59.2$ \\
\bottomrule
\end{tabular}
}
\caption{\small{}}
\label{tab:uavhumanrgb1}
\end{subtable}
\hspace{10pt}
\begin{subtable}{0.4\textwidth}
\centering
\resizebox{0.6\columnwidth}{!}{
\begin{tabular}{c c c c}
\toprule
FO & FA & Sampling & Top-1   \\
\midrule
\xmark & \xmark & \xmark & $21.06$ \\
\cmark & \xmark & \xmark & $25.89$ \\
\xmark & \cmark & \xmark & $24.15$ \\
\cmark & \cmark & \xmark & $27.00$ \\
\rowcolor{RowColorCode}
\cmark & \cmark & \cmark & $29.21$ \\
\bottomrule
\end{tabular}
}
\caption{\small{}}
\label{tab:uavhumanrgb_ablations}

\end{subtable}
\vspace{-40pt}
\end{table}
\begin{figure*}[t]
    \centering
    \captionsetup[subfigure]{labelformat=empty, font=tiny}
    \begin{subfigure}[b]{0.1\textwidth}
    \includegraphics[scale=0.09]{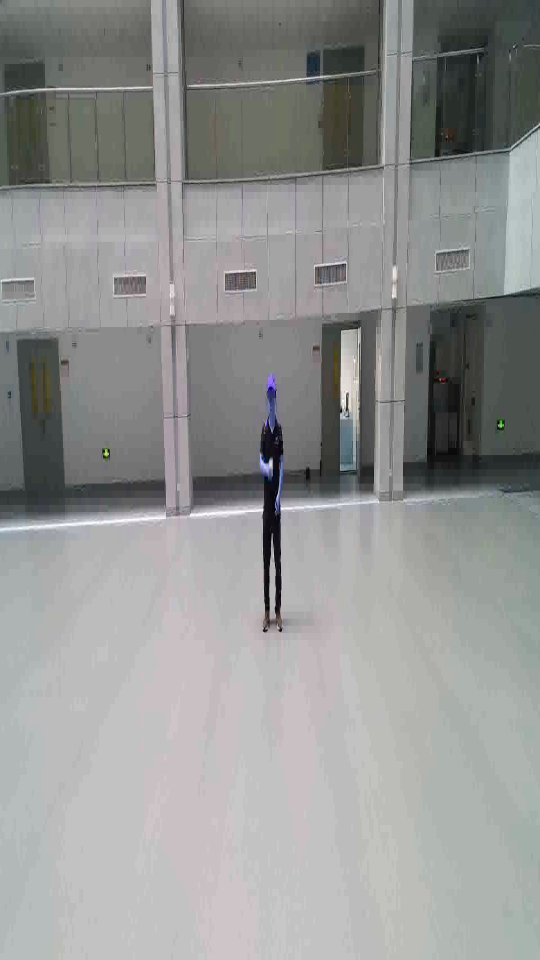}
    \caption{Drink}
    \end{subfigure}
    \begin{subfigure}[b]{0.1\textwidth}
    \includegraphics[scale=0.72]{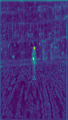}
    \caption{Before FO}
    \end{subfigure}
    \begin{subfigure}[b]{0.1\textwidth}
    \includegraphics[scale=0.72]{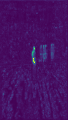}
    \caption{After FO}
    \end{subfigure}
    \begin{subfigure}[b]{0.1\textwidth}
    \includegraphics[scale=0.09]{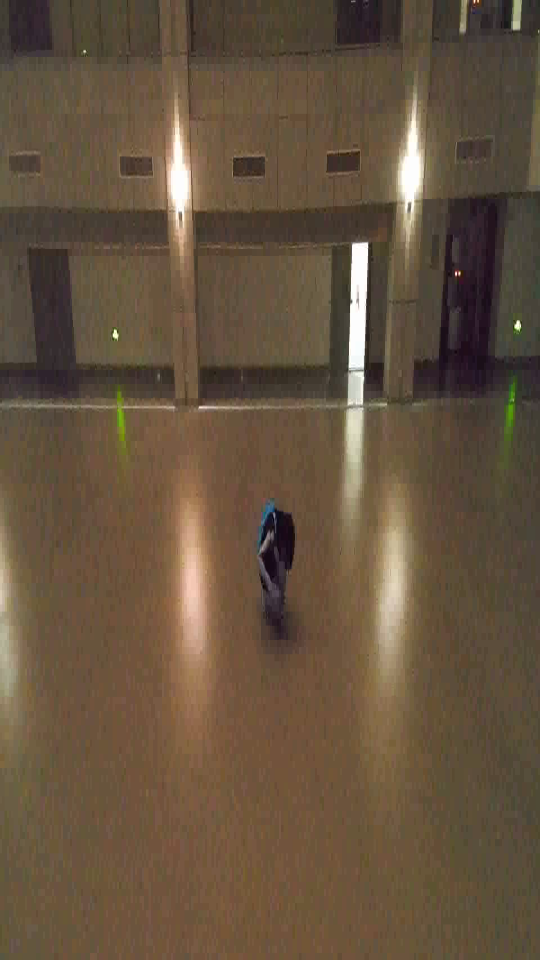}
    \caption{Dig a hole}
    \end{subfigure}
    \begin{subfigure}[b]{0.1\textwidth}
    \includegraphics[scale=0.72]{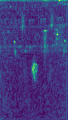}
    \caption{Before FO}
    \end{subfigure}
    \begin{subfigure}[b]{0.1\textwidth}
    \includegraphics[scale=0.72]{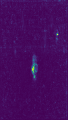}
    \caption{After FO}
    \end{subfigure}
    \begin{subfigure}[b]{0.1\textwidth}
    \includegraphics[scale=0.09]{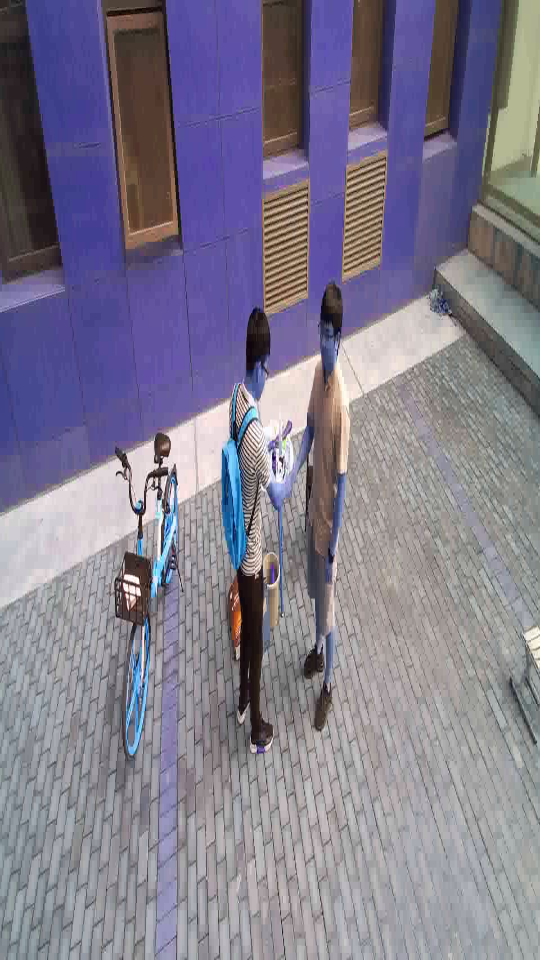}
    \caption{Shake hand}
    \end{subfigure}
    \begin{subfigure}[b]{0.1\textwidth}
    \includegraphics[scale=0.72]{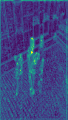}
    \caption{Before FO}
    \end{subfigure}
    \begin{subfigure}[b]{0.1\textwidth}
    \includegraphics[scale=0.72]{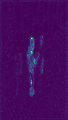}
    \caption{After FO}
    \end{subfigure}
    \\
    \begin{subfigure}[b]{0.1\textwidth}
    \includegraphics[scale=0.09]{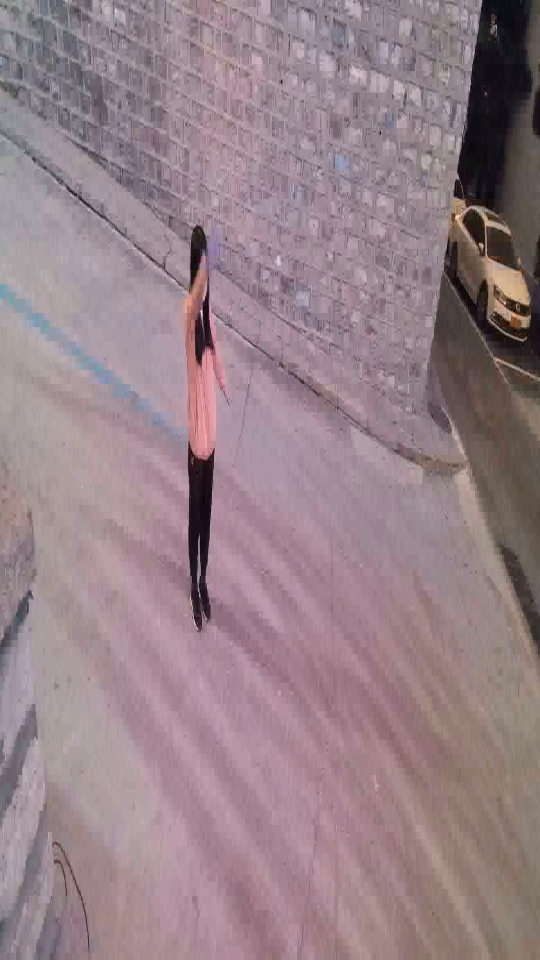}
    \caption{Left turn}
    \end{subfigure}
    \begin{subfigure}[b]{0.1\textwidth}
    \includegraphics[scale=0.72]{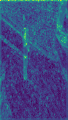}
    \caption{Before FO}
    \end{subfigure}
    \begin{subfigure}[b]{0.1\textwidth}
    \includegraphics[scale=0.72]{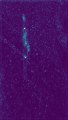}
    \caption{After FO}
    \end{subfigure}
    \begin{subfigure}[b]{0.1\textwidth}
    \includegraphics[scale=0.09]{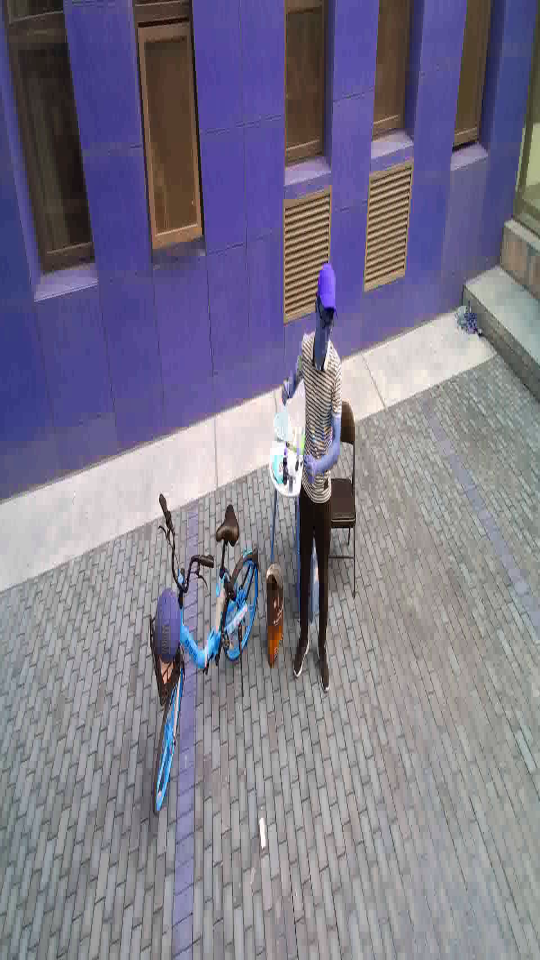}
    \caption{Punch }
    \end{subfigure}
    \begin{subfigure}[b]{0.1\textwidth}
    \includegraphics[scale=0.72]{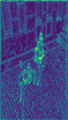}
    \caption{Before FO}
    \end{subfigure}
    \begin{subfigure}[b]{0.1\textwidth}
    \includegraphics[scale=0.72]{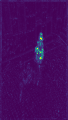}
    \caption{After FO}
    \end{subfigure}
    \begin{subfigure}[b]{0.1\textwidth}
    \includegraphics[scale=0.09]{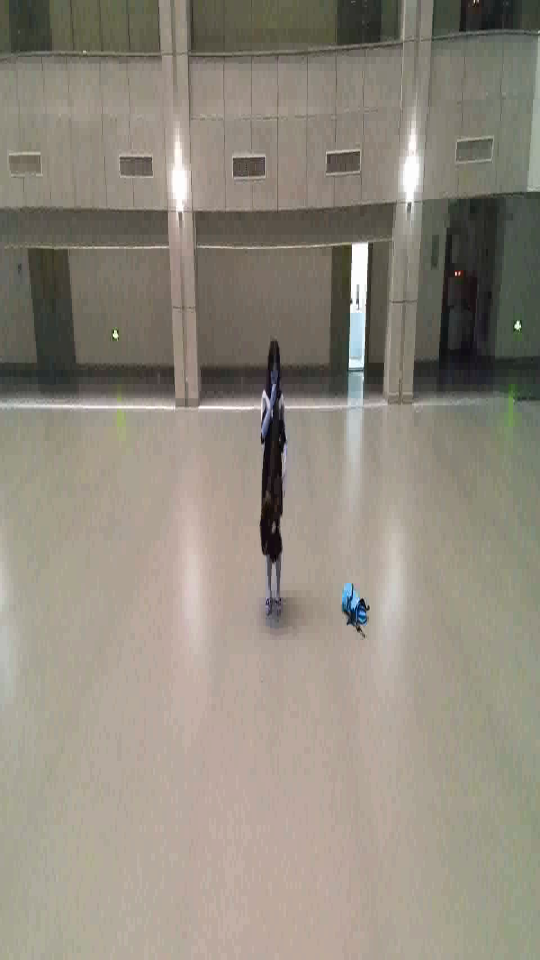}
    \caption{Rmv. coat}
    \end{subfigure}
    \begin{subfigure}[b]{0.1\textwidth}
    \includegraphics[scale=0.72]{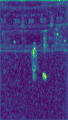}
    \caption{Before FO}
    \end{subfigure}
    \begin{subfigure}[b]{0.1\textwidth}
    \includegraphics[scale=0.72]{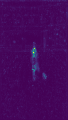}
    \caption{After FO}
    \end{subfigure}
   
    \caption{\small{\textbf{Qualitative results on UAV Human RGB.} We show the effect of our Fourier Object Disentanglement (FO) method. In each sample, the images, in order, correspond to a frame from the video, feature representation before disentanglement and the feature representation after disentanglement respectively. Notice the effectiveness of FO in scenes with light noise (Row 1 Image 2, Row 2 Image 3), dim light (Row 1 Image 2), dynamic camera and dynamic background (Row 1 Image 1). Regions of the scene corresponding to moving human actor (or salient dynamic) are amplified most (solid yellow). Static background is completely suppressed (solid purple). Static salient regions are slightly amplified (e.g. lower body of human actor in Row 2 Image 3 - yellow), and dynamic backgrounds are suppressed to a great extent (pale yellow in Row 1 Image 1). We show more results in the supplementary material.}}
    \label{fig:visualisations_uavhumanrgb}
    \vspace{-15pt}
\end{figure*}

\begin{figure}
    \centering
    \includegraphics[width=0.3\columnwidth]{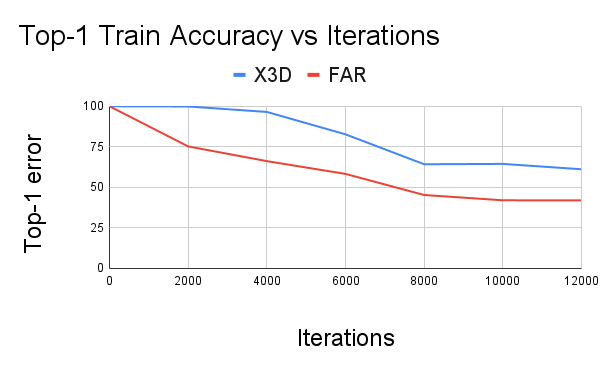}
    \includegraphics[width=0.3\columnwidth]{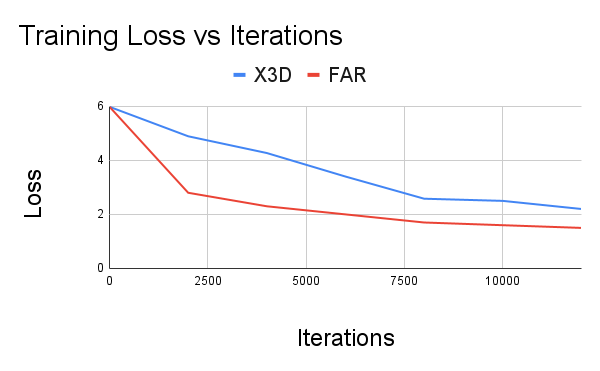}
    \caption{\small{\textbf{FAR converges much faster than the state-of-the-art action recognition method X3D-M \cite{feichtenhofer2020x3d}.} In the left curve, we show the top-1 train accuracy as a function of the networks' training iterations. In the right figure, we plot the training loss curve. We demonstrate that FAR imparts convergence benefits over prior work, under the same hyperparameter and GPU configurations.}}
    \label{fig:uavhumanrgb_convergence}
\end{figure}


\subsection{Main Results: UAV Human RGB}
\label{sec:results_uavhumanrgb}
\subsubsection{Benchmarking FAR}
\textit{FAR can be embedded within any 3D action recognition backbone to achieve state-of-the-art performance.} In Table \ref{tab:uavhumanrgb1}, we show results on UAVHuman RGB at different frame rates, input sizes, backbone network architectures and pre-trained weights initialization. In experiment (i) and (ii), we use the I3D backbone, and initialize the network with pretrained Kinetics weights. Spatially, we downsample the input video by a factor of $2$, and sample $8$ frames per video. This configuration gives the network full access to the spatial portions of the video at every stage of training and testing. 
FAR imparts a relative improvement of {\bf 38.69\%} and {\bf 23.18\%} in top-1 and top-5 accuracy, respectively. 

In the subsequent experiments, we use X3D-M, as the backbone. Many vision-based papers crop the original video into small patches of resolution $224\times224$. We explore this in experiments (iii)-(vi) under two settings: without initializing with Kinetics pretrained weights, and by initializing with Kinetics pretrained weights. Concurrent with our intuition, initializing with Kinetics pretrained weights results in better performance than without initializing with Kinetics pretrained weights. In both cases, with small crop siz, FAR improves performance over the corresponding baselines by $2.2\% - 4.24\%$. At a resolution of $224 \times 224$, there is a slight decrease in Top-5 accuracy ($0.1\%$). 

Video action recognition is a global level task. Hence, it is important for the network to see a larger spatial region of the video to understand context, and get a better view of the human actor. Moreover, since the design of FAR is specifically inspired of challenges pertaining to object background separation, and context encoding, the margin of improvement imparted by FAR to the backbone architecture is larger when the crop size is higher. At a crop size of $540\times540$, FAR improves top-1 and top-5 accuracies by $5.46\%$ and $3.67\%$ respectively, over the corresponding baselines. 
\begin{table}
\vspace{-1pt}
\centering
\caption{\small{\textbf{Comparisons with state-of-the-art self-attention based transformer methods on UAV Human.} We initialize all our models with Kinetics pre-trained weights. We observe higher accuracy and computational benefits (up to {\bf 3x}) with FAR.
}}

\resizebox{0.75\columnwidth}{!}{
\begin{tabular}{c c c c c c}
\toprule
Method & Param & Top-1 & FLOPs  &  GPU & Inference \\
 & M & \% & GFlops/video &  GB/video & Time (sec)/video\\
\midrule 
\multicolumn{6}{c}{I. Baseline non-attention models} \\
\midrule
X3D-M & $3.8$ & $36.6$ & $14.39$ & $3$ & $0.08$ \\
I3D-M & $12$ & $21.06$ & $346.55$ & $10$ & $0.1$ \\
\midrule 
\multicolumn{6}{c}{II. Comparisons with the state-of-the-art Fourier-based method } \\
\midrule
I3D + FNet \cite{fnet} (2021) & $12$ & $24.39$ & $346.56$ & $10$ & $0.1$ \\
\rowcolor{RowColorCode}
I3D + FAR (Ours) & $12$ & $29.21$ & $346.6$ & $10$ & $0.2$ \\
\midrule 
\multicolumn{6}{c}{III. Comparison with the state-of-the-art efficient-attention method} \\
\midrule
I3D + Efficient Attention \cite{shen2021efficient} (WACV 2021) & $12$ & $21.13$ & $462$ & $13.3$ & $0.12$\\
\rowcolor{RowColorCode}
I3D + Fourier Attention (Ours) & $12$ & $24.15$ & $346.57$ & $10$ & $0.19$\\
\midrule 
\multicolumn{6}{c}{IV. Comparisons with the state-of-the-art self-attention based transformer methods} \\
\midrule
ViT-B-TimeSformer (ICML 2021) \cite{bertasius2021space} & $121.4$ & $33.9$ & $2380$ & $32$ & $0.27$ \\
MVIT (ICCV 2021) \cite{fan2021multiscale} & $36.6$ & $24.3$ & $70.8$ & $9$ & $0.16$\\
\rowcolor{RowColorCode}
X3D-M + FAR  (Ours)& $3.8$ & $38.6$ & $14.41$ & $3$ & $0.09$ \\
\bottomrule
\end{tabular}
}
\label{tab:uavhumanrgb_sota_new}
\vspace{-20pt}
\end{table}

\subsubsection{Ablation Experiments}
\noindent \textbf{FAR ablations.} We present ablation experiments on the components of FAR in Table \ref{tab:uavhumanrgb_ablations}. We use the I3D backbone \cite{carreira2017quo}, and sample $8$ frames per video. We initialize with Kinetics pretrained weights, and spatially downsample (and then feed in the entire frame) the video by a factor of $2$. In the first four experiments, we uniformly sample $8$ frames from frame $0$. The first row is the baseline experiment with neither Fourier object disentanglement nor Fourier attention. We observe in the experiment corresponding to Row $2$ that object disentanglement improves performance by $22.9\%$ over the baseline. FO is a high pass filter (L2). The usage of a linear (L1) high pass filter results in an accuracy of $25.56\%$. Thus, we used the L2 high-pass filter as it results in a higher accuracy. Next, we determine the importance of context and long-range space-time relationships by using only Fourier attention, and demonstrate an improvement of $14.67\%$ in Row $3$. 

FO and FA complement each other - the former disentangles object from the background, while the latter decodes contextual information and inter-pixel, inter-frame relations. Using FO and FA in parallel, and sum-fusing the resultant feature maps cumulatively improves performance by $28.2\%$ over the baseline. Finally, we incorporate the sampling scheme, vis-a-vis, randomly initialized uniform sampling along with FO and FA. This results in a final accuracy of $29.21\%$, $38.69\%$ over the baseline.

\noindent \textbf{FA ablations.} We conduct ablation experiments on our proposed Fourier Attention (FA). We use the I3D backbone, and a video resolution of $540 \times 960 \times 8$. In all these experiments, FO is applied on level 2. In the first experiment, we extend FA to channels \cite{fu2019dual} in addition to space-time, at level $2$, the accuracy is $26.48$. In the second experiment, we apply channel FA at level $4$ \cite{fu2019dual} while retaining space-time FA at level $2$, the accuracy further degrades to $25.77$. In contrast, the accuracy with space-time Fourier attention is $27.00$. Thus, we find that global mixing at channel level does not contribute to improvement in performance. 

Next, we explore the notion of multi-level FA, where FA is applied at multiple levels and not just level $2$, the accuracy is $29.16$. In contrast, FAR's accuracy is $29.21$. Our conclusion is that FA extracts knowledge prerequisite to learning long-range space-time relationships at level $2$, to its maximum capacity, and applying it at more layers is redundant. 

\subsubsection{State-of-the-art comparisons} 

We report state-of-the-art comparisons in Table \ref{tab:uavhumanrgb_sota_new}. For all our experiments, we set the temporal and spatial resolutions at $8$ frames and upto $540$ (short side) respectively. We establish the baseline accuracies using non-attention networks vis-a-vis, I3D \cite{carreira2017quo} and X3D \cite{feichtenhofer2020x3d}, in experiment I. 

\noindent \textbf{Comparisons with FNet.} In Table \ref{tab:uavhumanrgb_sota_new} experiment II, we report the accuracy using FNet, which is the state-of-the-art Fourier transform based self-attention method. FNet naively replaces every self-attention layer with the Fourier transform of the feature maps at that level. The motivation is to "mix" different parts of the feature representation and thus gain global information. Originally designed for NLP, it achieves $92-97\%$ of the accuracy of BERT counterparts on the GLUE counterparts. However, when applied to video activity recognition on UAV Human RGB with the I3D backbone, we find that the accuracy is just $24.39\%$. In contrast, with the same backbone and hyperparameter settings, we demonstrate that FAR achieves $29.21\%$, an improvement of $19.76\%$. 

\noindent \textbf{Comparisons with efficient attention methods.} We compare with the current state-of-the-art efficient attention method \cite{shen2021efficient} in experiment III. For fair comparisons, we use the I3D backbone in both cases, at a video resolution of $540 \times 960 \times 8$. We demonstrate better accuracies with our Fourier Attention formulation at lower FLOPs and GPU memory. 

\noindent \textbf{Comparisons with transformer/self-attention methods.} We compare against self-attention based transformer methods in Table \ref{tab:uavhumanrgb_sota_new}. Specifically, we compare against (i) TimesFormer \cite{bertasius2021space} (ICML 2021) - a self-attention video recognition method based on joint space-time self attention, and (ii) MViT (ICCV 2021) - a transformer based method that combines multi-scale feature hierarchies. We demonstrate much better performance at lower number of FLOPs, GPU memory and inference time. Another benefit is that FAR does not add any new parameters to the neural network and uses the same number of parameters as its backbone network. In contrast, MVIT and TimeSformer use much higher number of parameters.

\begin{table}
\vspace{-20pt}
\centering
\caption{\small{\textbf{Results on more UAV datasets}. We demonstrate that FAR improves the state-of-the-art accuracy by \textbf{8.02\%-17.61\%} across popular UAV datasets. }}
\resizebox{0.4\columnwidth}{!}{
\begin{tabular}{c c c c c}
\toprule
Method & Frames & Input Size & Init. & Top-1   \\
\midrule
\multicolumn{5}{c}{\textbf{(i) Dataset: UAV Human Night \cite{li2021uav}}} \\
\midrule 
I3D \cite{carreira2017quo}& $8$ & $480\times640$ & Kinetics & $28.72$ \\
\rowcolor{RowColorCode}
FAR & $8$ & $480\times640$ & Kinetics & $33.78$ \\
\midrule 
\multicolumn{5}{c}{\textbf{(ii) Dataset: Drone Action \cite{perera2019drone}}} \\
\midrule 
HLPF \cite{jhuang2013towards}&  All & $1920\times1080$ & None & $64.36$ \\
PCNN \cite{cheron2015p} &  - & $1920\times1080$ & None & $75.92$\\
X3D-M \cite{feichtenhofer2020x3d}&  $16$ & $224\times224$ & Kinetics & $83.4$ \\
\rowcolor{RowColorCode}
FAR &  $16$ & $224\times224$ & Kinetics & $92.7$ \\
\midrule 
\multicolumn{5}{c}{\textbf{(iii) NEC Drone \cite{choi2020unsupervised}}}\\
\midrule 
X3D-M \cite{feichtenhofer2020x3d} &  $8$ & $960\times 540$ & Kinetics & $66.15$ \\
\rowcolor{RowColorCode}
FAR &  $8$ & $960\times 540$ & Kinetics & $71.46$ \\

\bottomrule
\end{tabular}
}

\label{tab:more_drones}
\vspace{-20pt}
\end{table}


\subsection{Results: More UAV Datasets}
\label{sec:results_moreuav} 
We demonstrate the effectiveness of FAR on multiple UAV benchmarks in Table \ref{tab:more_drones}. We demonstrate that FAR outperforms prior work by \textbf{17.61\%, 11.15\% and 8.02\%} on UAV Human Night, Drone Action, and NEC Drone respectively. 



\section{Conclusions, Limitations and Future Work}

We present a new method for UAV Video Action Recognition. Our method exploits the mathematical properties of Fourier transform to automatically disentangle object from the background, and to encode long-range space-time relationships in a computationally efficient manner. We demonstrate benefits in terms of accuracy, computational complexity and training time on multiple UAV datasets. Our method has a few limitations. 
The sampling strategy based on randomly initialization is a naive method to span all video frames. It might be interesting to explore the usage of better video sampling strategies \cite{zhi2021mgsampler,korbar2019scsampler}. 
Next, we assume that the input videos contain only one human agent performing action. Multi-action videos could be a potential extension of our method. Moreover, we believe that FAR can be extended to other tasks such as video object segmentation and video generation, front-camera action recognition, graphics and rendering \cite{lloyd2008logarithmic,mitchell1988reconstruction}. 

\noindent \textbf{Acknowledgements:} We thank Rohan Chandra for reviewing the paper. This research has been supported by ARO Grants W911NF1910069, W911NF2110026  and Army Cooperative Agreement W911NF2120076.

\section*{A. Appendix}

\subsection*{A.1. Datasets}

We describe the UAV datasets used for evaluating FAR. 

\paragraph{UAV Human RGB~\cite{li2021uav}:} UAV Human is the largest UAV-based human behavior understanding dataset. Split $1$ contains $15172$ and $5556$ images for training and testing respectively. This challenging dataset covers human actions captured under varying illumination, time of day (daytime, nighttime), different subjects and backgrounds, weathers, occlusions, etc, across $155$ diverse human actions. UAV Human RGB is collected by drones with an Azure Kinect DK camera. The videos are of resolution $1920\times1080$. The dataset is available at  https://sutdcv.github.io/uav-human-web/. 

\paragraph{UAV Human Night Camera~\cite{li2021uav}:} UAV Human Night Camera contains videos similar to UAV Human RGB captured using a night-vision camera. The night vision camera captures videos in color mode in the daytime, and grey-scale mode in the nighttime. The resolution of the videos is $640\times480$. The dataset is available at  https://sutdcv.github.io/uav-human-web/.

\paragraph{Drone Action~\cite{perera2019drone}:} Drone Action is an outdoor drone video dataset captured using a free flying drone. It has $240$ HD RGB videos with $66919$ frames, across $13$ human actions. The dataset is available at https://asankagp.github.io/droneaction/.

\paragraph{NEC Drone~\cite{choi2020unsupervised}:} NEC Drone dataset is an indoor UAV video dataset with $16$ human actions captured by a DJI Phantom 4.0 pro v2 drone, performed by human subjects in an unconstrained manner. The dataset contains $2079$ labeled videos at a resolution of $1920\times1080$. It has $10$ single person actions such as walk, run, jump, etc, and $6$ two person actions such as shake hands, push a person, etc. The dataset is available at https://www.nec-labs.com/~mas/NEC-Drone/. 

\subsection*{A.2. Implementation Details}

In the interest of reproducibility, we will make all code and pretrained models publicly available upon acceptance of the paper. We also attach the codes used in our experiments with the supplementary zip folder submitted for review. 

\label{exp:implementation}
\paragraph{Backbone network architecture:} We benchmark our models using two state-of-the-art video recognition backbone architectures (i) I3D \cite{carreira2017quo} (CVPR 2017) (ii) X3D-M \cite{feichtenhofer2020x3d} (CVPR 2020). I3D is a 3D inflated CNN, based on 2D CNN inflation, and enables the learning of spatial-temporal features. X3D is also a 3D inflated CNN, and progressively expands a 2D CNN along multiple network axes such as space, time, width and depth. 
For both X3D and I3D, we extract mid-level features after the second layer.

\paragraph{Training details:} Our models were trained using NVIDIA GeForce 1080 Ti GPUs, and NVIDIA RTX A5000 GPUs. Initial learning rates were \{$0.01$, and $0.001$\} across datasets. We use cosine annealing and poly annealing for learning rate decay in X3D and I3D respectively, 
We use the Stochastic Gradient Descent (SGD) optimizer with weight decay of $0.0005$ and momentum of $0.9$, and cosine/ poly annealing for learning rate decay. The final softmax predictions of all our models were constrained using multi-class cross entropy loss. 

\subsection*{A.3. Fourier Disentanglement}

Videos depicting human action have four types of entities: moving salient regions (typically corresponding to moving object), static salient regions (typically corresponding to static object), moving non-salient regions (typically corresponding to dynamic background), and static non-salient regions (typically corresponding to static background). Robust action recognition systems should learn features that heavily amplify moving objects, followed by static objects (that provide contextual cues and are relevant to the prediction). This should be followed by background entities. According to our formulation, dynamic salient regions are amplified the most. This is because the Fourier mask highlights dynamic regions, and the features learnt by the network have a higher amplitude at the salient regions. Static non-salient regions are at the other end of the spectrum because the Fourier mask suppresses these regions, as well as the features learnt by the network have a lower amplitude at the non-salient regions. Static-salient and dynamic salient regions lie at the middle of the spectrum. The final equation for Fourier disentanglement uses the $l2$ operation in the computation of $M_{FO}$ and linear application of $f$. This implies that static salient regions have a higher amplitude than the dynamic non-salient regions. Thus, the ordering of amplitudes that is formed as: dynamic-salient $>$ static-salient $>$ dynamic-non-salient $>$ static-non-salient, in concordance with the relevance for decision making for action recognition. Thus, static as well as dynamic background regions have lower amplitudes than static and dynamic regions of the object executing action. 

In addition, the video may contain noise (light noise or otherwise) and camera movement. In regions of the video where there is noise, the amplitude of the feature map depicting saliency will be low. Hence, noise gets suppressed. Any movement of non-salient pixels due to camera motion gets suppressed since they are a part of dynamic non-salient regions. Moreover, camera motion is generally uniform across the spatial dimensions of the video (covering salient as well as non-salient regions). Thus, it doesn't impact the decision making ability of the aerial video recognition system.  

\noindent \textbf{Comparisons with motion-based methods.} Motion-based methods either model spatial and temporal information separately using two-stream 2D CNNs \cite{lee2018motion} or use motion representation as an auxiliary guiding factor to 3D CNNs. The latter is very expensive \cite{piergiovanni2019representation}. In contrast, we jointly model space and time using a 3D backbone, and then disentangle the moving human actor from the background using FO. Prior work has demonstrated the superiority \cite{feichtenhofer2019slowfast,feichtenhofer2020x3d} of 3D CNNs over two-stream 2D CNNs. FO imparts a relative improvement of $22.93\%$ over the 3D I3D backbone and can be used with any 3D CNN to achieve state-of-the-art performance.

\subsection*{A.4. Fourier Attention}

\begin{lemma}
Given an input matrix A, Fourier attention as well self-attention \cite{vaswani2017attention,bertasius2021space} encapsulate long-range relationships for global mixing by computing outer products.
\end{lemma}
\paragraph{Proof}
\textbf{Self-attention:} Without loss of generality, let $[a_{ij}]$ denote the elements of a square matrix A (with dimensions $N$) in $2D$. $f$, $g$, $h$ represent $1\times 1$ convolutions for key, query, value computations in self-attention. Hence, key, query and value vectors are $[fa_{ij}]$, $[ga_{ij}]$ and $[ha_{ij}]$ respectively. The first step of self-attention is the computation of sub-attention, which is the matrix multiplication of the transpose of query with key, which is $[ga_{ij}]^{T} \odot [fa_{ij}]$, which is equal to $\sum_{i=1}^{N} ga_{mi} \times fa_{in}$. The next step is the computation of self-attention, which is the matrix multiplication of the value vector with the transpose of sub-attention, which is equal to $[ha_{ij}] \odot \sum_{k=1}^{N} ga_{lk} \times fa_{kn}$. Hence, the self-attention matrix $S_{mn}$ is:
\begin{equation}
    S_{mn} = \sum_{l=1}^{N} ha_{ml} \sum_{k=1}^{N} [ ga_{lk} \times fa_{kn}]
    \label{eq:sa}
\end{equation}
\textbf{Fourier-attention:} Without loss of generality, let $[a_{ij}]$ denote the elements of a square matrix A (with dimensions $N$) in $2D$. The Fourier transform is $\sum_{i=1}^{N} \sum_{j=1}^{N} \exp(\minus 2\pi mi/N) \exp(\minus 2\pi nj/N)$. Multiplication of the Fourier transform with its conjugate transpose, and inverse FFT gives us $$\sum_{b=1}^{N} \sum_{c=1}^{N} \exp(\minus 2\pi mc/N \minus 2\pi nb/N)a_{mn}  \times \nonumber  \{\sum_{j=1}^{N} \sum_{i=1}^{N} \exp(\minus 2\pi j(b\minus c)/N \minus 2\pi i(c\minus b)/N)a_{ij}^2\}$$. Finally, weighted multiplication of the above term with $[a_{ij}]$ and a careful rearrangement of the terms involved leads us to the final expression for Fourier attention. Fourier attention $F_{mn}$ is:
\begin{align}
    F_{mn} = \sum_{b=1}^{N} \sum_{c=1}^{N} \overbrace{\exp(\minus 2\pi mc/N) \exp(\minus 2\pi nb/N)}^{h_{mn}(b,c)}a_{mn}  \times \nonumber \\[-10pt] \{\sum_{j=1}^{N} \sum_{i=1}^{N} \underbrace{\exp(\minus 2\pi j(b\minus c)/N)}_{f_{mn}(b,c)} a_{ij} \times \underbrace{\exp(\minus 2\pi i(c\minus b)/N)}_{g_{mn}(b,c)}a_{ij}\}
    \label{eq:fa}
\end{align}

$f$, $g$, $h$ in Equation \ref{eq:sa} are $1\times 1$ convolutions, and that the exponential terms span the entire spectrum of frequencies lets us define $f$, $g$, $h$ for Fourier attention as shown in Equation \ref{eq:fa}. Thus, the equation for Fourier attention can be simplified as:
\begin{align}
    F_{mn} = \sum_{b=1}^{N} \sum_{c=1}^{N} h_{mn}(b,c)a_{mn} \times \nonumber \\[-10pt] \{\sum_{j=1}^{N} \sum_{i=1}^{N} f_{mn}(b,c)a_{ij} \times g_{mn}(b,c)a_{ij}\}
    \label{eq:fa1}
\end{align}
In self-attention, f,g,h are learnable. In contrast, in Fourier attention, f,g,h are pre-defined by the Fourier spectrum. Nonetheless, they exhaustively cover the Fourier spectrum. Moreover, the terms involved and the structure of computations (multiplications followed by summation) in Equations \ref{eq:sa} and \ref{eq:fa1} are similar, both promote global mixing and encapsulate long-range relationships. 

\subsection{A.5. Future Work: Extension to Multi-Agent Systems}

We mainly focus on popular UAV datasets that consist of single human agent performing an action to validate our Fourier object disentanglement (FO) method. We plan to extend our method to multi-agent systems as a part of future work. Our formulation of FO should work for multi-agent systems. Corresponding to the regions with multiple human actors (which are all dynamic salient regions), the value of $F_{FO}$ will be high, the equations described in Section 3 for FO will remain unchanged. Thus, FO can disentangle multiple human actors in the scene without any external bounding boxes. This is because the formulation based on frequency of pixels and saliency activations highlights any region (even for multiple actors) in the video that has \textit{salient dynamic} objects i.e. actors performing action. This is done intrinsically, within the computation of the networks' feature maps. 

For {\bf multi-agent systems}, the system needs to (spatially) localize the human actor, in addition to classifying the action of each actor. To do this, action localization systems [82] such as typically setup an object detection-like pipeline with bounding boxes regressors and classifiers. Just as our FO method can be embedded within any 3D backbone (such as SlowFast or [82] or I3D or X3D) for improved action classification (Section 5.3), our FO method can also be embedded within any 3D backbone for improved action localization. The region highlights inferred by FO corresponding to pixels with multiple human actors will help the downstream bounding box regression as well as classification modules perform better in multi-agent scenes. 
\begin{figure*}[t]
    \centering
    \captionsetup[subfigure]{labelformat=empty, font=tiny}
    \begin{subfigure}[b]{0.1\textwidth}
    \includegraphics[scale=0.09]{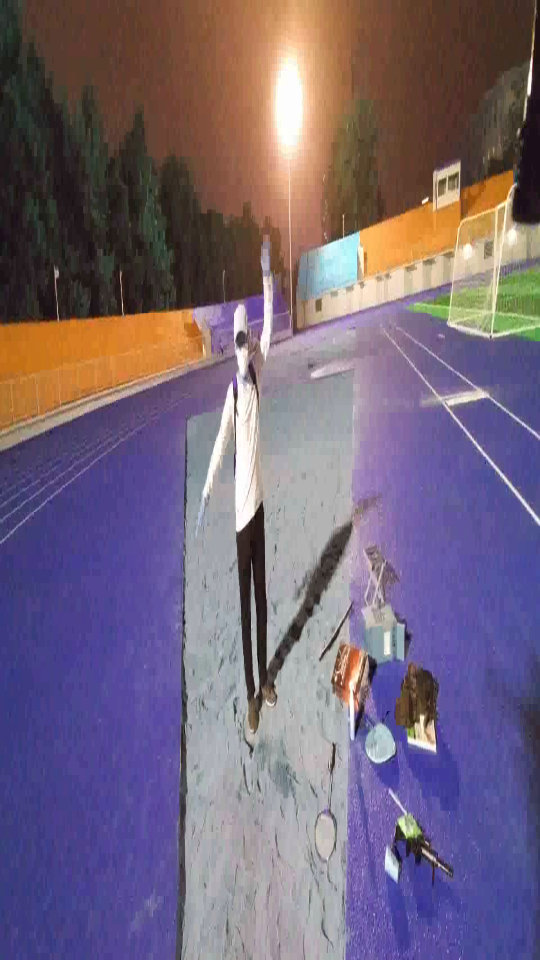}
    \caption{rear rt.turn}
    \end{subfigure}
    \begin{subfigure}[b]{0.1\textwidth}
    \includegraphics[scale=0.72]{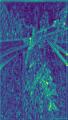}
    \caption{Before FO}
    \end{subfigure}
    \begin{subfigure}[b]{0.1\textwidth}
    \includegraphics[scale=0.72]{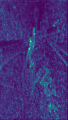}
    \caption{After FO}
    \end{subfigure}
    \begin{subfigure}[b]{0.1\textwidth}
    \includegraphics[scale=0.09]{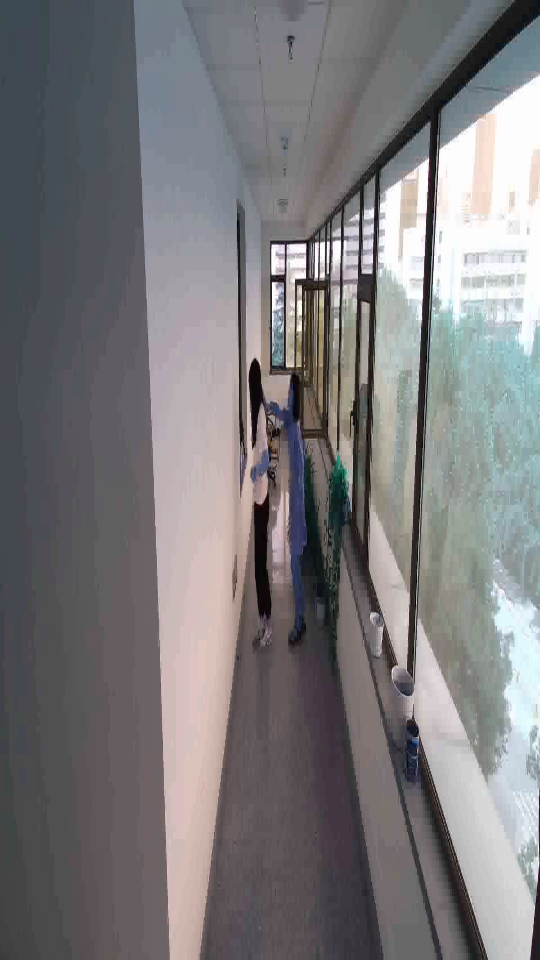}
    \caption{chaseHumn}
    \end{subfigure}
    \begin{subfigure}[b]{0.1\textwidth}
    \includegraphics[scale=0.72]{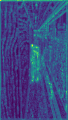}
    \caption{Before FO}
    \end{subfigure}
    \begin{subfigure}[b]{0.1\textwidth}
    \includegraphics[scale=0.72]{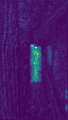}
    \caption{After FO}
    \end{subfigure}
    \begin{subfigure}[b]{0.1\textwidth}
    \includegraphics[scale=0.09]{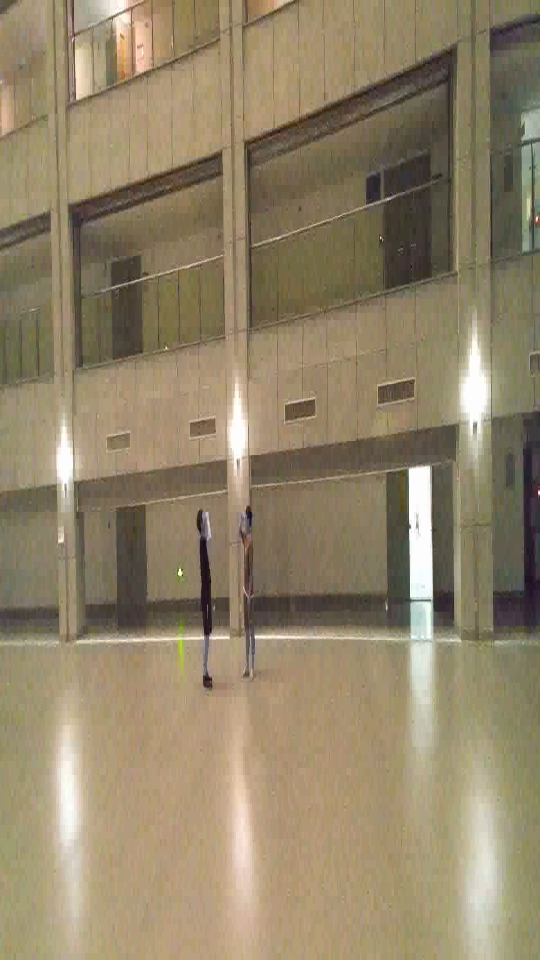}
    \caption{Drink toast}
    \end{subfigure}
    \begin{subfigure}[b]{0.1\textwidth}
    \includegraphics[scale=0.72]{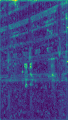}
    \caption{Before FO}
    \end{subfigure}
    \begin{subfigure}[b]{0.1\textwidth}
    \includegraphics[scale=0.72]{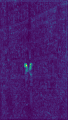}
    \caption{After FO}
    \end{subfigure}
    \\
    \begin{subfigure}[b]{0.1\textwidth}
    \includegraphics[scale=0.09]{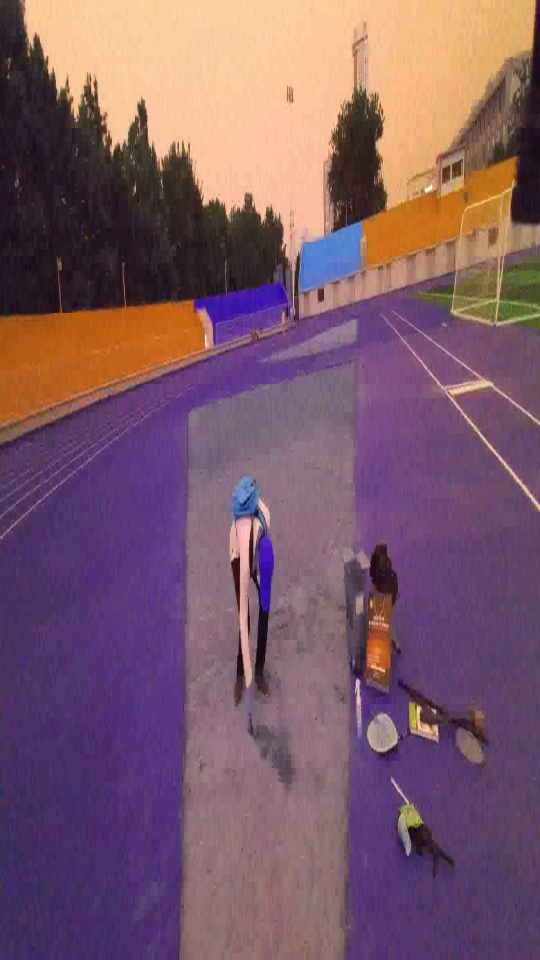}
    \caption{Dig a hole}
    \end{subfigure}
    \begin{subfigure}[b]{0.1\textwidth}
    \includegraphics[scale=0.72]{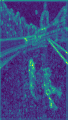}
    \caption{Before FO}
    \end{subfigure}
    \begin{subfigure}[b]{0.1\textwidth}
    \includegraphics[scale=0.72]{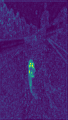}
    \caption{After FO}
    \end{subfigure}
    \begin{subfigure}[b]{0.1\textwidth}
    \includegraphics[scale=0.09]{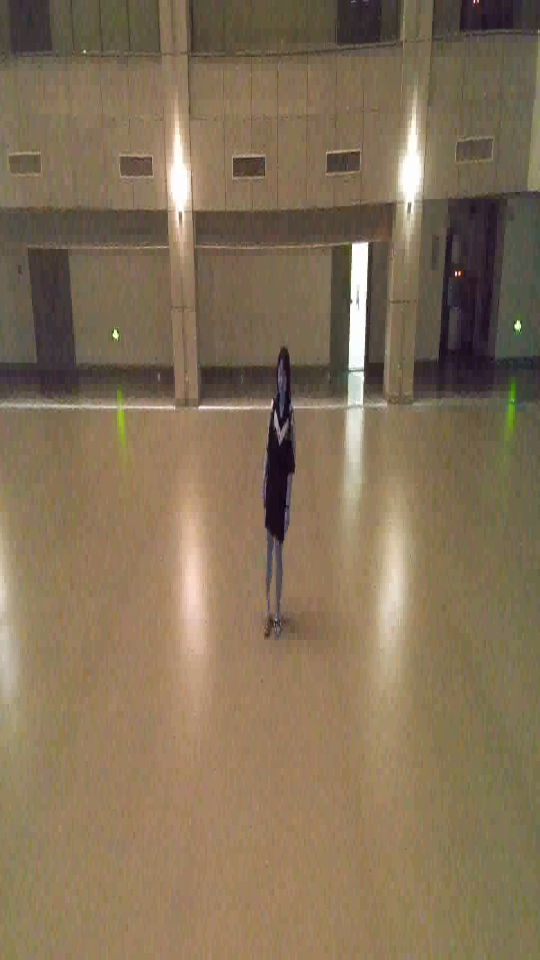}
    \caption{Kick aside}
    \end{subfigure}
    \begin{subfigure}[b]{0.1\textwidth}
    \includegraphics[scale=0.72]{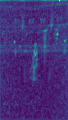}
    \caption{Before FO}
    \end{subfigure}
    \begin{subfigure}[b]{0.1\textwidth}
    \includegraphics[scale=0.72]{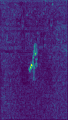}
    \caption{After FO}
    \end{subfigure}
    \begin{subfigure}[b]{0.1\textwidth}
    \includegraphics[scale=0.09]{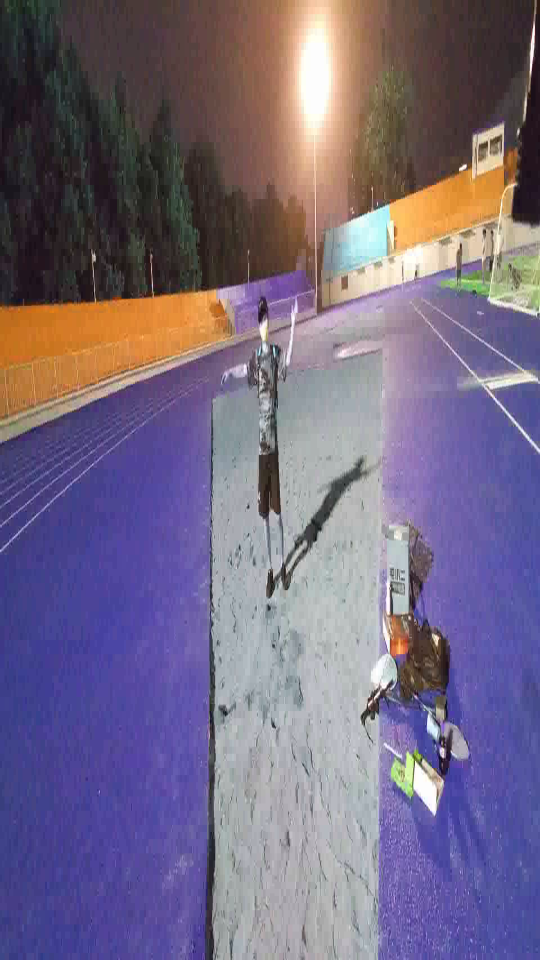}
    \caption{Move left}
    \end{subfigure}
    \begin{subfigure}[b]{0.1\textwidth}
    \includegraphics[scale=0.72]{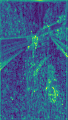}
    \caption{Before FO}
    \end{subfigure}
    \begin{subfigure}[b]{0.1\textwidth}
    \includegraphics[scale=0.72]{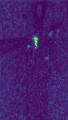}
    \caption{After FO}
    \end{subfigure}
   
    \caption{\small{\textbf{Qualitative results on UAV Human RGB.} We show the effect of our Fourier Object Disentanglement (FO) method. In each sample, the images, in order, correspond to a frame from the video, feature representation before disentanglement and the feature representation after disentanglement respectively. Notice the effectiveness of FO in scenes with light noise, dim light, dynamic camera and dynamic background. Regions of the scene corresponding to moving human actor (or salient dynamic) are amplified most (solid yellow). Static background is completely suppressed (solid purple). Static salient regions are slightly amplified, and dynamic backgrounds are suppressed to a great extent. We show videos depicting various complexities along with the predictions in the video file attached with the supplementary.}}
    \label{fig:visualisations_uavhumanrgb1}
    
\end{figure*}

\begin{figure*}[t]
    \centering
    \begin{subfigure}[b]{0.24\textwidth}
    \includegraphics[scale=0.2]{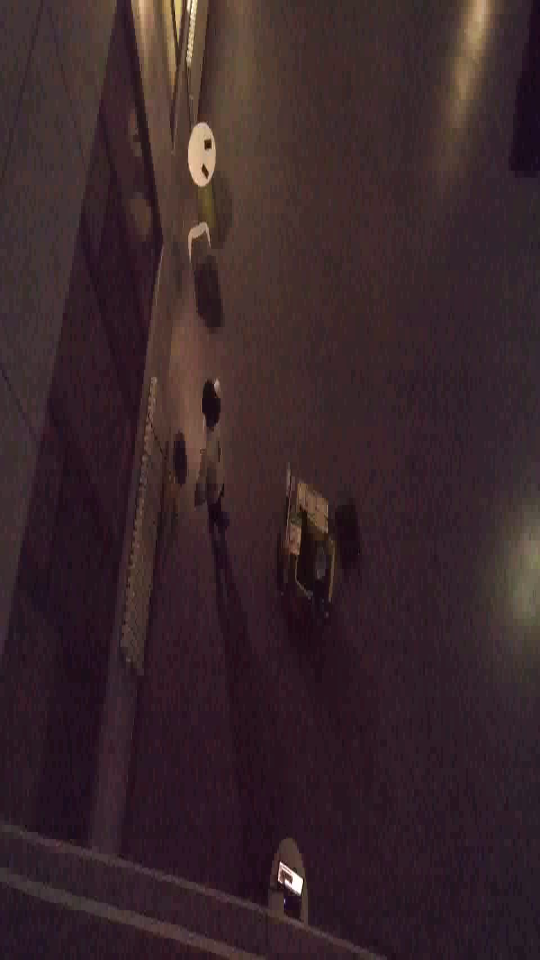}
    \caption{Predicted: Drop something \\ GT: Put hands on hips}
    \end{subfigure}
    \begin{subfigure}[b]{0.24\textwidth}
    \includegraphics[scale=0.2]{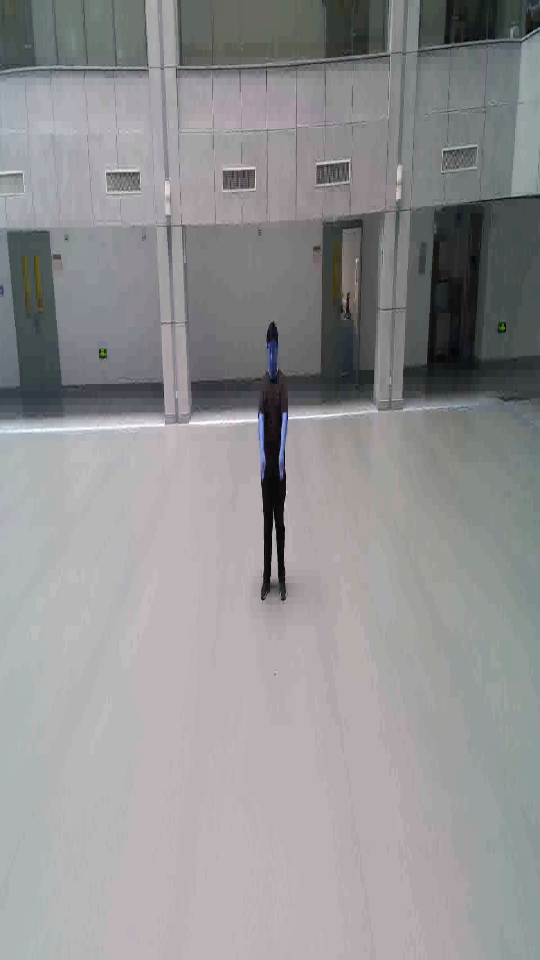}
    \caption{Predicted: Open the bottle \\ GT: Decelerate \\}
    \end{subfigure}
    \begin{subfigure}[b]{0.24\textwidth}
    \includegraphics[scale=0.2]{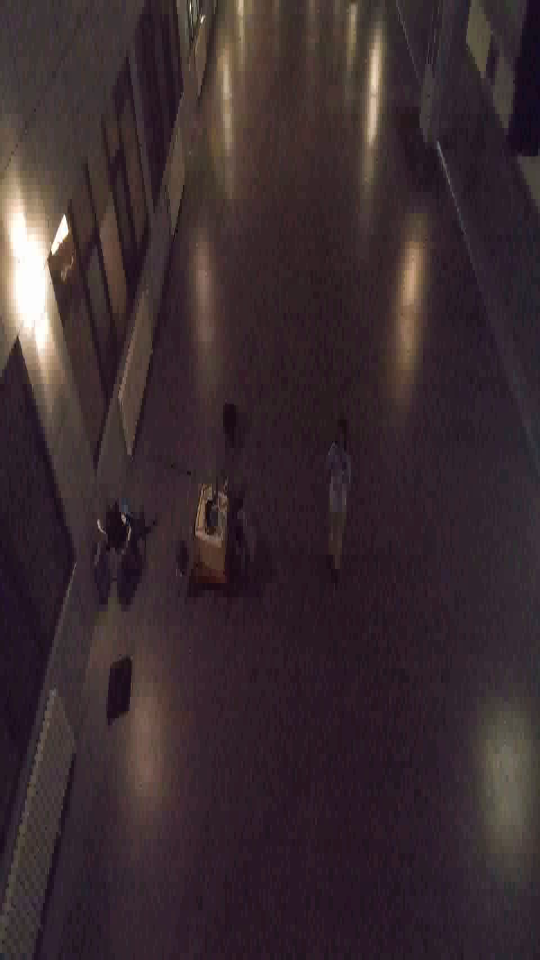}
    \caption{Predicted: Punch with fists \\ GT: Cheer \\}
    \end{subfigure}
    \begin{subfigure}[b]{0.24\textwidth}
    \includegraphics[scale=0.2]{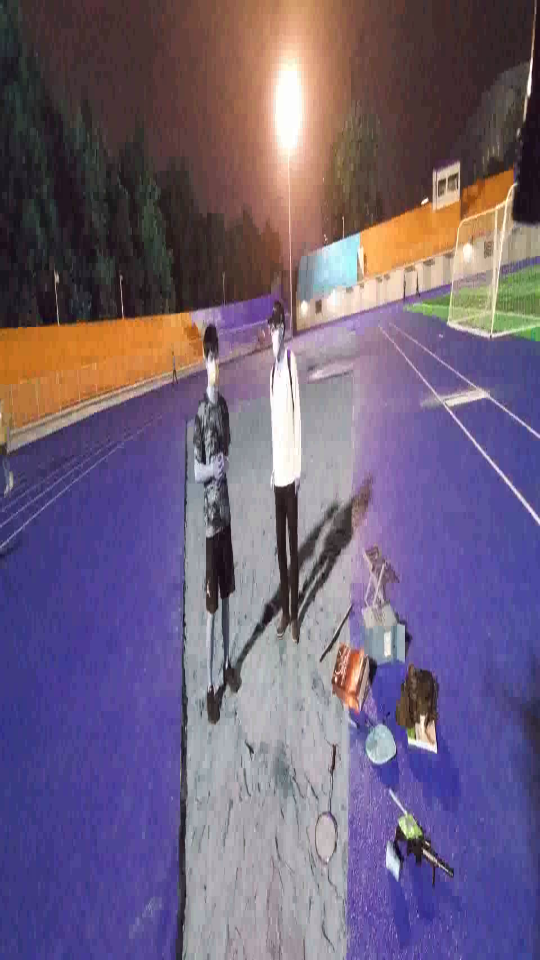}
    \caption{Predicted: Pushing someone \\ GT: Rob something from someone}
    \end{subfigure}\\
    
    \begin{subfigure}[b]{0.24\textwidth}
    \includegraphics[scale=0.2]{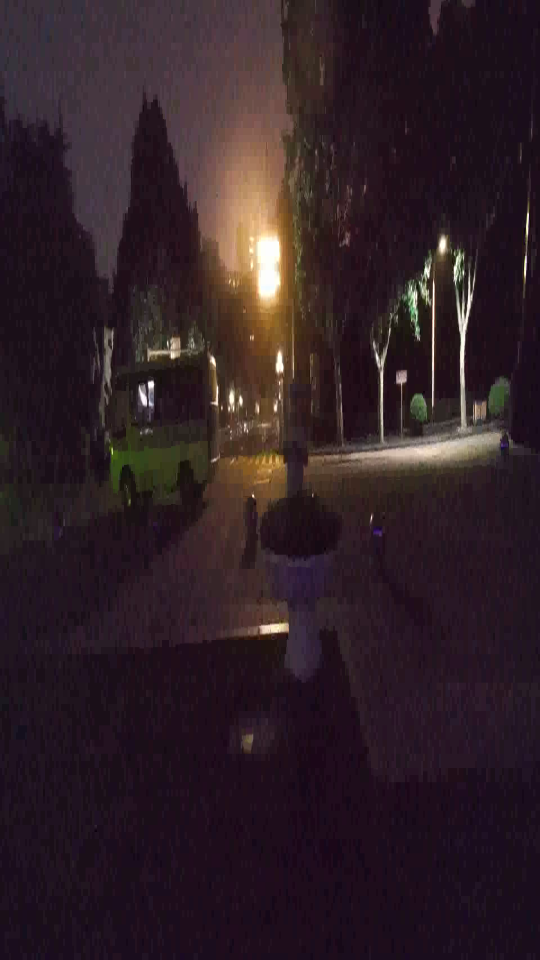}
    \caption{Predicted: Smoke \\ \\ GT: Apply cream to hands}
    \end{subfigure}
    \begin{subfigure}[b]{0.24\textwidth}
    \includegraphics[scale=0.2]{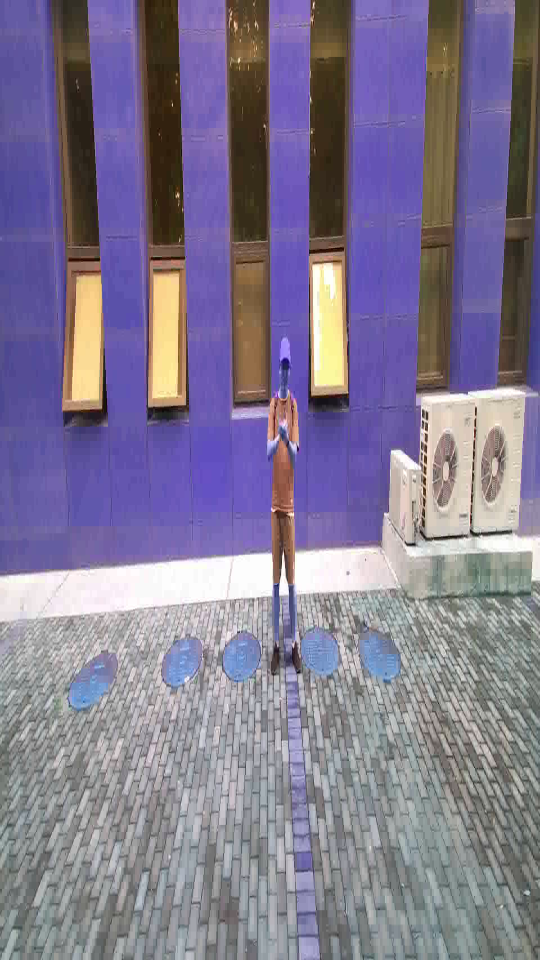}
    \caption{Predicted: Play with cell phones \\ GT: Applaud \\}
    \end{subfigure}
    \begin{subfigure}[b]{0.24\textwidth}
    \includegraphics[scale=0.2]{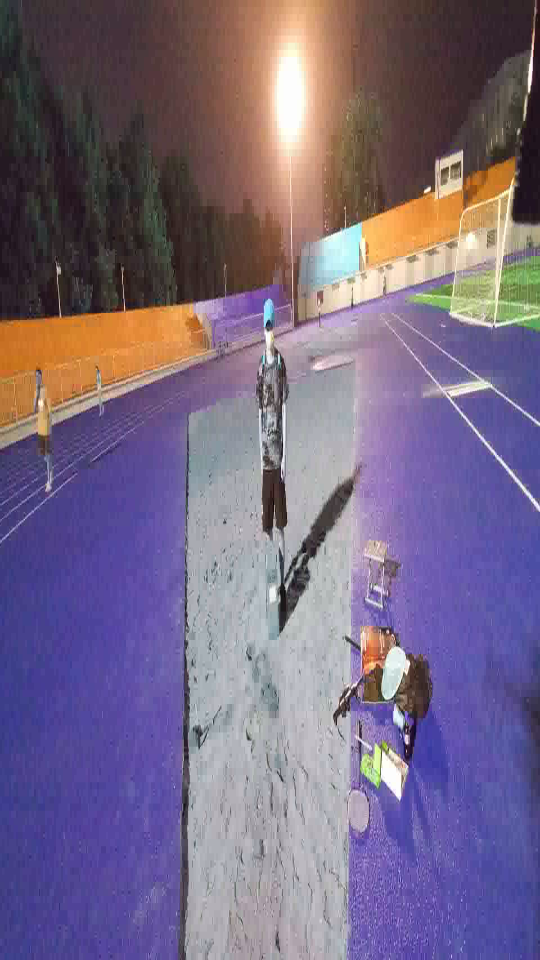}
    \caption{Predicted: Blow nose \\ GT: Throw litter \\}
    \end{subfigure}
    \begin{subfigure}[b]{0.24\textwidth}
    \includegraphics[scale=0.2]{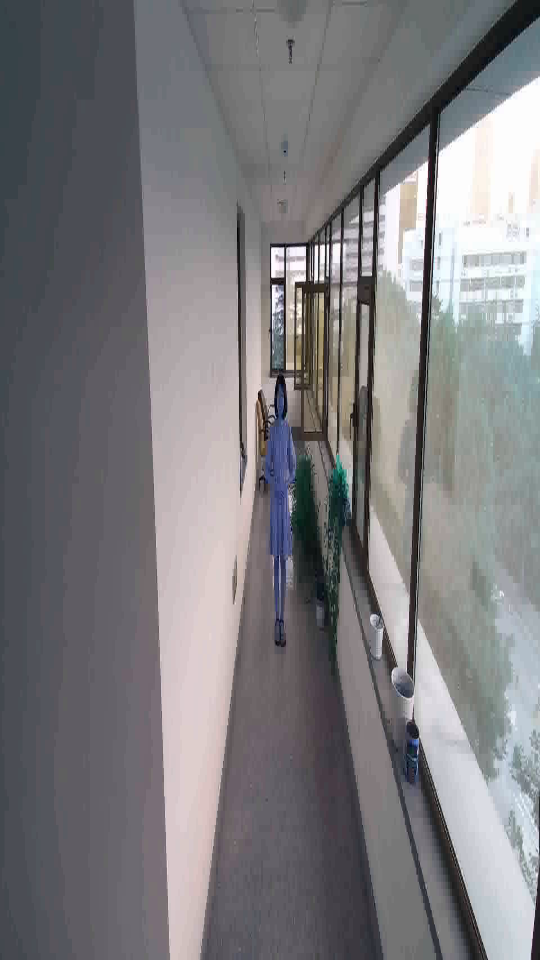}
    \caption{Predicted: Applaud \\ GT: Cross palms together}
    \end{subfigure}
    \caption{\textbf{Failure cases on UAV Human RGB.} We show frames from UAV Human RGB videos where \model~predicts the wrong class. In many cases, we observe that the predicted class has pixel level interactions similar to the ground truth. For instance, in case (d), both, predicted class and GT are two-person actions, and entail one person harming the other. Similarly, in video (h), both actions involve interaction between the two hands of a person. In video (a), both actions correspond to a human standing straight with hands at hip level. It would be interesting to explore learning distinguishable feature representations for the $155$ classes as a part of future work.}
    \label{fig:fail_uavhumanrgb}
    
\end{figure*}

\clearpage

\bibliographystyle{splncs04}
\bibliography{references}
\end{document}